\documentclass[runningheads]{llncs}

\usepackage{accv}

\usepackage{accvabbrv}

\usepackage{graphicx}
\usepackage{booktabs}
\usepackage{amsmath}
\usepackage{amssymb}

\usepackage{graphbox}

\usepackage{makecell}
\usepackage{tabularx}
\usepackage{multirow}
\usepackage{multicol}
\usepackage{diagbox}  
\usepackage{threeparttable}

\usepackage[T1]{fontenc}    
\usepackage{amsfonts}       
\usepackage{pifont}
\usepackage{dsfont}

\usepackage{algorithm}
\usepackage{algorithmic}

\usepackage{color}
\usepackage[dvipsnames,svgnames,table]{xcolor}

\usepackage[utf8]{inputenc} 
\usepackage{url}            
\usepackage{mathtools}
\usepackage{nicefrac}       
\usepackage{tikz}
\usetikzlibrary{calc}

\usepackage[normalem]{ulem}

\usepackage[accsupp]{axessibility}  

\usepackage[pagebackref,breaklinks,colorlinks,citecolor=accvblue]{hyperref}

\usepackage{orcidlink}

\begin{document}

\long\def\ignorethis#1{}


\newcommand{\img}[1]{\mathbf{I}_{\text{#1}}}
\newcommand{\paren}[1]{\left( #1 \right)}
\newcommand{\bparen}[1]{\left[ #1 \right]}
\newcommand{\feature}[1]{\phi \paren{#1}}
\newcommand{\normtwo}[1]{\lVert #1 \rVert_2^2}
\newcommand{\normone}[1]{\left\lVert #1 \right\rVert_1}

\newcommand{\Paragraph}[1]{\vspace{1mm}\noindent\textbf{#1}}

\newcommand{\figref}[1]{Figure~\cref{fig:#1}}
\newcommand{\tabref}[1]{Table~\ref{tab:#1}} 
\newcommand{\itmref}[1]{[\ref{itm:#1}]}     
\newcommand{\eqnref}[1]{\eqref{eq:#1}}
\newcommand{\secref}[1]{Section~\ref{sec:#1}}
\newcommand{\eqmain}[1]{(\textcolor{blue}{#1})}
\newcommand{\fakeref}[1]{\textcolor{accvblue}{#1}}
\newcommand{\fakeeqref}[1]{(\textcolor{red}{#1})}
\newcommand{\fakecite}[1]{[\textcolor{accvblue}{#1}]}

\newcommand{\mb}[1]{\mathbf{#1}}
\newcommand{\bs}[1]{\boldsymbol{#1}}
\newcommand{\n}{\mbox{\qquad}}              
\newcommand{\red}[1]{{\color{red}#1}}

\newcommand{\ignore}[1]{}   
\newcommand{\cmt}[1]{\begin{sloppypar}\large\textcolor{red}{#1}\end{sloppypar}}

\newcommand{\TODO}[1]{\textcolor{red}{[TODO]\{#1\}}}
\newcommand{\todo}[1]{\textcolor{red}{#1}}
\newcommand{\torevise}[1]{\textcolor{blue}{#1}}
\newcommand{\revise}[1]{\textcolor{blue}{#1}}
\newcommand{\copied}[1]{\textcolor{red}{[COPIED: #1]}}

\newcommand{\jaeha}[1]{{\textcolor{BurntOrange}{\textbf{Jaeha: }#1}}}
\newcommand{\sanghyun}[1]{{\textcolor{blue}{\textbf{Sanghyun: }#1}}}
\newcommand{\seungjun}[1]{{\textcolor{ProcessBlue}{\textbf{Seungjun: }#1}}}
\newcommand{\kyoungmu}[1]{{\textcolor{OrangeRed}{\textbf{Kyoung Mu: }#1}}}
\newcommand{\needref}{[\textcolor{blue}{put}, \textcolor{blue}{some}, \textcolor{blue}{references}]}

\newcommand{\tabspace}{\vspace{-2mm}}
\newcommand{\tabxspace}{\vspace{-4mm}}
\newcommand{\figspace}{\vspace{-2mm}}
\newcommand{\figxspace}{\vspace{-3mm}}
\newcommand{\best}[1]{{\textcolor{red}{\textbf{#1}}}}
\newcommand{\secondbest}[1]{{\textcolor{blue}{\underline{#1}}}}

\newcommand{\set}[1]{\{#1\}}

\def\PE{\Phi}

\newcommand{\mpage}[2]
{
\begin{minipage}{#1\linewidth}\centering
#2
\end{minipage}
}

\newcommand{\citenumber}[1]{[\textcolor{cvprblue}{#1}]}
\definecolor{BurntOrange}{rgb}{0.8,0.33,0.0}
\definecolor{DarkBlue}{rgb}{0.0,0.0,0.75}

\newcolumntype{L}[1]{>{\raggedright\let\newline\\\arraybackslash\hspace{0pt}}m{#1}}
\newcolumntype{C}[1]{>{\centering\let\newline\\\arraybackslash\hspace{0pt}}m{#1}}
\newcolumntype{R}[1]{>{\raggedleft\let\newline\\\arraybackslash\hspace{0pt}}m{#1}}

\definecolor{green4mark}{RGB}{21, 152, 56}
\definecolor{red4mark}{RGB}{252, 54, 65}
\definecolor{orange4mark}{RGB}{247, 135, 2}
\newcommand{\cmark}{\textcolor{green4mark}{\ding{51}}}
\newcommand{\xmark}{\textcolor{red4mark}{\ding{55}}}
\def\grayP{15}
\def\cyanP{20}
\def\greenP{18}
\def\limeP{15}
\def\SpringGreenP{12}
\newcommand{\splitT}{0.255}
\title{Noise-Free One-Step LoRA for Task-Driven Image Restoration with Diffusion Priors}

\titlerunning{Noise-Free One-Step LoRA for Task-Driven Image Restoration}

\author{Jaeha Kim\inst{1} \and
Kyoung Mu Lee\inst{1,2}}

\authorrunning{J.~Kim and K.M.~Lee}

\institute{Dept.\ of ECE \& ASRI, Seoul National University, Seoul, Korea \and
IPAI, Seoul National University, Seoul, Korea\\
\email{jhkim97s2@gmail.com, kyoungmu@snu.ac.kr}}

\maketitle

\begin{abstract}

Degraded images not only reduce visual quality but also impair downstream high-level vision tasks.
Task-driven image restoration (TDIR) addresses this issue by jointly optimizing restoration quality and task performance.
Recent works show that pretrained diffusion priors benefit TDIR, yet diffusion-based restoration is inherently stochastic, as the sampling process depends on a random noise term, which can undermine task consistency.
In this paper, we show that a deterministic, noise-free one-step forward pass with pretrained diffusion priors can substantially improve TDIR, but the benefit critically depends on the adaptation module: LoRA yields consistent gains, whereas ControlNet-style conditioning does not.
This enables one-step forwarding that surpasses conventional multi-step diffusion TDIR baselines.
Furthermore, we introduce a task-preserving GAN training strategy that improves perceptual quality without sacrificing task performance.
Extensive experiments on classification, segmentation, and detection demonstrate consistent gains over prior TDIR methods, and we further validate generalization on real-world degraded images and OCR.

\end{abstract}
\section{Introduction}
\label{sec:intro}

Image restoration~(IR) is a fundamental computer vision problem that aims to reconstruct high-quality~(HQ) images from degraded low-quality~(LQ) observations.
Image degradation harms visual quality, and IR has therefore been developed primarily to maximize visual fidelity for human perception, as measured by distortion- or perception-based metrics~\cite{wu2023qalign, ke2021musiq, yang2022maniqa}.

However, in many practical scenarios, images are consumed not only by humans but also by downstream high-level vision systems, such as image classification, semantic segmentation, object detection, and optical character recognition.
In such cases, image degradation impairs not only visual quality but also the performance of these systems.
Crucially, prior studies~\cite{dai2016image, pei2018does} reveal that classic IR methods, which aim to maximize visual fidelity, do not necessarily recover the task performance lost to degradation.
This limitation has motivated task-driven image restoration~(TDIR), whose goal is to restore degraded images that are both visually plausible and beneficial for a target high-level vision task.
To achieve this, TDIR methods~\cite{sr_tdsr, liu2017image, son2020urie, huang2020dsnet} jointly optimize the IR model together with the task network, rather than training the IR model independently, thereby improving downstream task performance over classic IR.

\begin{figure*}[t!]
    \centering
    \captionsetup[subfigure]{labelfont=scriptsize, textfont=scriptsize}

    \begin{minipage}{\textwidth}
        \centering
        \subfloat[HQ (Ground-truth)]{\includegraphics[width=0.24\linewidth]{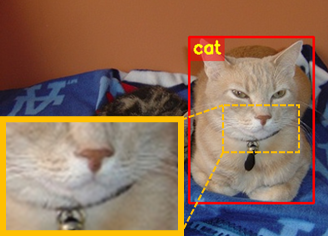}}
        \hfill
        \subfloat[LQ]{\includegraphics[width=0.24\linewidth]{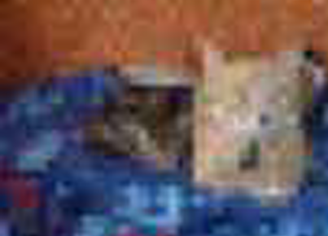}}
        \hfill
        \subfloat[EDTR (run 1)]{\includegraphics[width=0.24\linewidth]{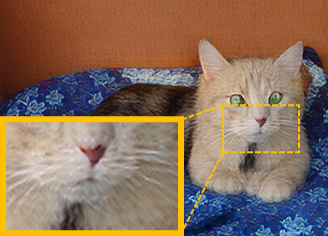}}
        \hfill
        \subfloat[EDTR (run 2)]{\includegraphics[width=0.24\linewidth]{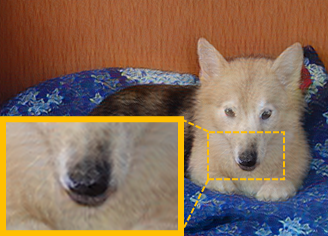}}
    \end{minipage}%

    \caption{
    Stochasticity in diffusion-based TDIR.
    Given the same LQ input, EDTR~\cite{kim2025exploiting} can produce different restorations across sampling runs, leading to inconsistent recovery of task-relevant details (\eg, (c) recovers the cat mouth/whiskers, while (d) introduces a dog-like nose pattern).
    (a) shows the HQ image with detection ground-truth boxes, and (b) the degraded LQ input, while (c--d) show restored images only.
}
    \label{fig:stochasticy-tdir}
\vspace{-3mm}
\end{figure*}

Recently, TDIR methods~\cite{kim2025exploiting, chen2025unirestore} have incorporated pretrained diffusion priors such as Stable Diffusion~(SD)~\cite{rombach2022high}, achieving large downstream task gains by reconstructing semantically meaningful content even when the input evidence is incomplete.
However, diffusion-based restoration is inherently \emph{stochastic}: the reverse-denoising process depends on a random noise term~\cite{ho2020denoising, song2020denoising}, so the same LQ input yields different restorations across runs.
As shown in \cref{fig:stochasticy-tdir}, this causes fluctuations in task-relevant details and undermines prediction consistency---a particularly harmful property for TDIR, which is far more sensitive to such details than perceptual IR.
Moreover, these methods typically rely on multi-step sampling, incurring inference cost that is unfavorable for downstream pipelines.

In this work, we ask whether the stochasticity of diffusion priors can be avoided while retaining their benefits for TDIR.
To this end, we replace the stochastic multi-step sampling~\cite{ho2020denoising, song2020denoising} with a single deterministic forward pass that feeds the LQ input directly to the pretrained denoiser without any random noise term, yielding a noise-free one-step restoration.
Interestingly, the benefit of this design hinges on the adaptation module: it yields substantial gains with LoRA~\cite{hu2022lora}, but fails to improve---and can even degrade---task performance with ControlNet~\cite{zhang2023adding}-style conditioning.
We attribute this contrast to an input-domain shift induced by noise-free inputs (relative to SD pretraining on noisy latents), and provide evidence for it: under the noise-free regime, LoRA recovers task-relevant details more faithfully in the task-feature space and is more robust to input perturbations, whereas ControlNet is not.
We term this approach NOLA (Noise-Free One-Step LoRA Adaptation) and refer to the resulting restoration network as NOLA-IR.
Although one-step diffusion with LoRA has been explored for perceptual IR~\cite{wu2024one}, such methods neither target downstream tasks nor analyze how stochasticity affects them; we are, to our knowledge, the first to study noise-free forwarding from a task-driven perspective and to show that its benefit critically depends on the adaptation module.

While NOLA's single-step restoration improves task performance, it can yield lower perceptual quality than multi-step sampling.
Naively adding an adversarial objective~\cite{gans} improves realism but conflicts with task-driven optimization and degrades downstream performance.
We therefore introduce task-preserving GAN training, which regularizes adversarial fine-tuning to stay close to the task-driven restoration in task-relevant details, improving visual quality without sacrificing task performance.

We extensively validate NOLA-IR not as an isolated benchmark method but as a practical TDIR system.
On classification, segmentation, and detection, it surpasses prior TDIR baselines in both task performance and perceptual quality, and remains effective on real-world degraded images.
We also provide optical character recognition (OCR) experiments to demonstrate the generality of our approach beyond standard high-level vision tasks.

We summarize our contributions as follows:

\begin{itemize}
\item We show that stochasticity when leveraging pretrained diffusion priors can undermine TDIR, and propose NOLA, a deterministic noise-free one-step forward pass with LoRA adaptation. We further demonstrate that the benefit of the noise-free forward pass depends on the adaptation module.
\item We introduce task-preserving GAN training to improve perceptual quality without sacrificing task performance.
\item Extensive experiments on classification, segmentation, and detection demonstrate consistent gains over prior diffusion-based TDIR baselines, and we further validate generalization on real-world LQ images and OCR.
\end{itemize}

\section{Related Works}
\label{sec:related-works}

\subsection{Perceptual Image Restoration}

Early IR methods~\cite{sr_edsr, db_nafnet, sr_swinir, sr_rdn, db_restormer, sr_srcnn, sr_vdsr} optimized distortion-based objectives (\eg, PSNR/SSIM~\cite{measure_ssim}), but higher PSNR does not necessarily align with human preference~\cite{zhang2018unreasonable}, driving the community toward perceptual IR~\cite{sr_srgan, sr_esrgan, rs_realesrgan, zhang2021designing} with generative models and human-aligned metrics~\cite{wu2023qalign, measure_niqe, ke2021musiq, yang2022maniqa, heusel2017gans, wang2023exploring}.
With diffusion models (\eg, DDPM~\cite{ho2020denoising}), perceptual IR began incorporating diffusion priors: DiffBIR~\cite{lin2023diffbir} adapts SD~\cite{rombach2022high} via IRControlNet, a ControlNet~\cite{zhang2023adding}-style module tailored to IR, and subsequent works~\cite{yu2024scaling, wu2024seesr, yang2023pasd, sun2024coser} exploit richer language-vision priors~\cite{liu2023llava, zhang2024recognize, li2023blip}.
Another line of research~\cite{wu2024one, wang2024sinsr, dong2025tsd, yue2024efficient, lin2025harnessing} focuses on accelerating diffusion-based IR by reducing the number of sampling steps.

Among accelerating approaches, OSEDiff~\cite{wu2024one} shares the one-step, noise-free inference setting with our approach.
However, it targets perceptual IR and neither examines downstream task performance nor analyzes how stochasticity in diffusion-based restoration affects it.
In contrast, we study noise-free forwarding from a downstream-task perspective, analyze the impact of stochasticity on TDIR, and systematically investigate how the adaptation module (\eg, ControlNet vs.\ LoRA) governs this behavior.

\subsection{Task-driven Image Restoration}

Unlike perceptual IR, task-driven image restoration (TDIR) focuses on improving downstream task performance under image degradations by training the IR network with task-specific supervision.
Early approaches~\cite{liu2017image, liu2022image}, including TDSR~\cite{sr_tdsr} and URIE~\cite{son2020urie}, directly incorporate task losses (\eg, classification/segmentation/detection objectives) during restoration training, demonstrating improved downstream performance.
Subsequent works~\cite{liu2020connecting, yang2023visual, li2023detection, qin2024scene}, including RSRSSN~\cite{zhao2018residual} and SR4IR~\cite{kim2024beyond}, further explore feature-level alignment by optimizing restorations in the deep feature space of the task network.

Recently, TDIR has also incorporated pretrained diffusion models (\eg, SD) as strong generative priors.
UniRestore~\cite{chen2025unirestore} adapts diffusion features from SD for downstream tasks, demonstrating that diffusion priors can be effective for both perceptual quality and downstream task performance.
EDTR~\cite{kim2025exploiting} further observes that excessive generative behavior of diffusion priors can be detrimental to task performance, and proposes to initialize from mildly noisy latents with short-step denoising to modulate the prior toward restoring task-relevant details.
However, such multi-step diffusion-based TDIR methods still rely on stochastic sampling, which we identify as a key obstacle for task consistency and address through deterministic noise-free forwarding.

\section{Method}
\label{sec:method}

\subsection{Problem Definition}
\label{ssec:problem-definition}

Let $\{(x^{\mathrm{LQ}}_i, x^{\mathrm{HQ}}_i, y_i)\}_{i=1}^{N}$ denote paired LQ and HQ images with the corresponding task label $y_i$ for a downstream high-level vision task (\eg, classification).
We consider a trainable restoration network $\mathcal{G}_{\theta}$ and a trainable high-level vision task network $\mathcal{H}_{\phi}$.
Our goal is to improve downstream task performance under degraded inputs by jointly training $\mathcal{G}_{\theta}$ and $\mathcal{H}_{\phi}$ with task-aware supervision, while also producing perceptually pleasing restorations.
We evaluate both downstream task performance and restoration quality.

\subsection{Noise-Free One-Step LoRA Adaptation}
\label{sec:method:nola}

\subsubsection{Motivation.}

Diffusion-based IR methods~\cite{lin2023diffbir, wang2024exploiting, wu2024seesr}, including diffusion-based TDIR methods~\cite{kim2025exploiting, chen2025unirestore}, commonly involve a random noise term in the sampling process~\cite{ho2020denoising, nichol2021improved}, so the same LQ input yields different restorations across runs.
While such variability may be tolerable for perceptual IR, TDIR is more sensitive: fluctuations in fine details can alter task-relevant cues and undermine prediction consistency.
As illustrated in \cref{fig:stochasticy-tdir}, one run recovers structures important for detection (\eg, the cat mouth/whiskers), whereas another introduces different fine-grained patterns (\eg, a dog-like nose), which can mislead the downstream task and make task-network optimization harder.

\begin{figure}[tb]
    \centering
    \captionsetup[subfigure]{labelfont=scriptsize, textfont=scriptsize}
    \renewcommand{\wp}{0.33}
    \resizebox{1.0\linewidth}{!}{
            \centering
            \subfloat{\includegraphics[width=\wp\linewidth]{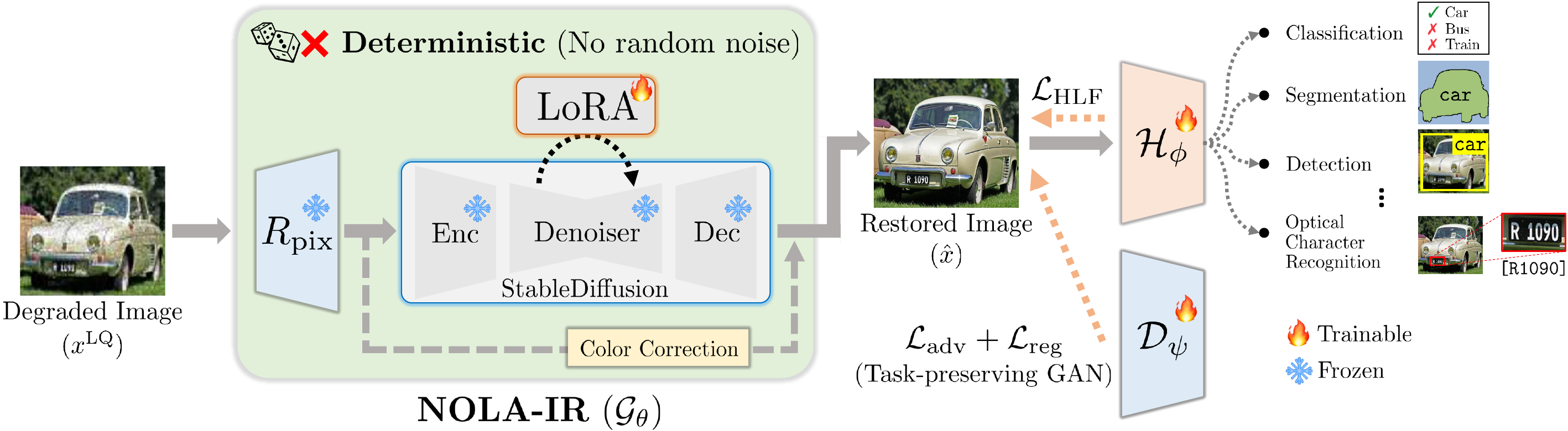}}
            \vspace{-2mm}
    }
    \caption{Overview of our task-driven image restoration. Given a degraded image, the restoration network $\mathcal{G}_\theta$ produces a restored image via a single deterministic forward pass (no random noise term). $\mathcal{G}_\theta$ is built on a frozen Stable Diffusion backbone, adapted with a lightweight LoRA module, and further refined with task-preserving GAN training. The restored image is then fed into a trainable task network $\mathcal{H}_\phi$ for downstream tasks.}
    \label{fig:nola}
    \vspace{-2mm}
\end{figure}

\vspace{-2mm}
\subsubsection{Noise-Free One-Step Forward Pass.}

Motivated by the above, we replace the stochastic sampling process with a single deterministic forward pass.
Concretely, we repurpose the pretrained diffusion prior as a restoration module by avoiding any random noise term in the forward computation.
Given an LQ image $x^{\mathrm{LQ}}$, we deterministically obtain the restored image by
\begin{equation}
    \hat{x} = \mathcal{G}_{\theta}(x^{\mathrm{LQ}}),
    \label{eq:one-step}
\end{equation}
where $\mathcal{G}_{\theta}$ is built on a frozen pretrained SD backbone with lightweight trainable adapter modules.
Following prior SD-based restoration work~\cite{lin2023diffbir, kim2025exploiting}, $\mathcal{G}_{\theta}$ also includes a pre-restoration module ($R_\text{pix}$) and wavelet-based color correction~\cite{wang2024exploiting} (see Supplementary \cref{sec:architecture_details} for details).
We note that, rather than proposing a new architecture, our contribution lies in identifying that a noise-free one-step forward pass, together with an appropriate adaptation module, is key to effective TDIR with pretrained diffusion priors.

\vspace{-2mm}
\subsubsection{Choice of Adaptation Module.}

Although \cref{eq:one-step} yields a deterministic forward pass, its impact on task performance depends on the adaptation module used to adapt the frozen SD backbone.
Empirically, LoRA~\cite{hu2022lora} yields consistent gains under \cref{eq:one-step}, whereas the same noise-free forward pass with ControlNet-style conditioning~\cite{lin2023diffbir} does not improve, and can even degrade, performance.

We attribute this contrast to an input-domain shift: removing the noise term drives the denoiser input away from the noisy-latent distribution seen during SD pretraining, and adapters differ in their capacity to cope with this shift.
ControlNet steers a frozen denoiser through additive conditioning, which is limited in compensating for this mismatch, whereas LoRA directly updates the denoiser weights and adapts more effectively under the noise-free regime.
We substantiate this explanation in \cref{ssec:abl-nola}, where we show that, under noise-free forwarding, LoRA recovers task-relevant details more faithfully in the task-feature space and is more robust to input perturbations than ControlNet.
Based on these observations, we build our method on LoRA and term it Noise-Free One-Step LoRA Adaptation (NOLA); we refer to the resulting IR network $\mathcal{G}_\theta$ as NOLA-IR.
\cref{fig:nola} provides an overview of the proposed pipeline.

\begin{figure}[tb]
    \centering
    \captionsetup[subfigure]{labelfont=scriptsize, textfont=scriptsize}
    \renewcommand{\wp}{0.33}
    \resizebox{0.75\linewidth}{!}{
            \centering
            \subfloat{\includegraphics[width=\wp\linewidth]{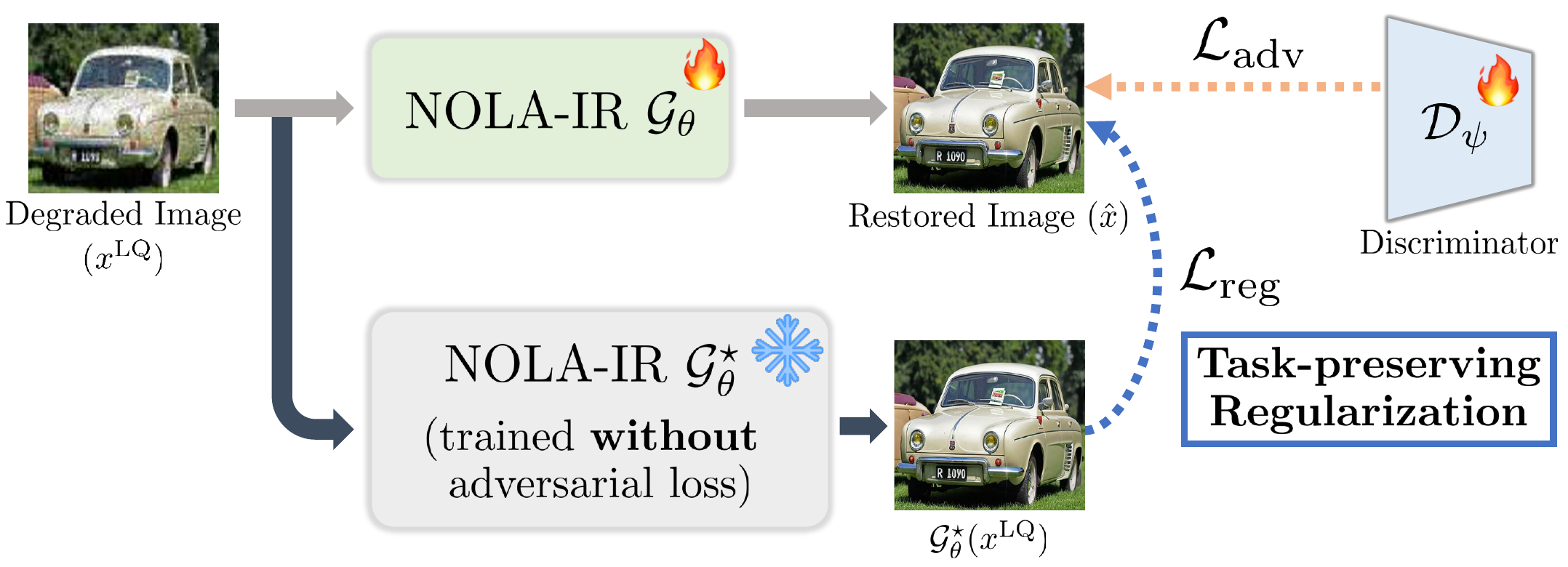}}
    }
    \vspace{-4mm}
    \caption{Illustration of task-preserving GAN training.
}
    \label{fig:tpgan}
    \vspace{-3mm}
\end{figure}

\subsection{Task-Preserving GAN Training}
\label{sec:method:tp-gan}

While NOLA improves task performance, its single-step restoration can yield lower perceptual quality than multi-step diffusion sampling.
To address this, we incorporate adversarial training~\cite{gans}.
However, we observe that naively adding an adversarial objective can conflict with task-driven optimization and consequently degrade downstream performance.

To balance perceptual realism and downstream task performance, we introduce a task-preserving regularization that reduces the conflict between adversarial and task-driven objectives during training.
We first train NOLA-IR without adversarial loss and denote it as $\mathcal{G}_{\theta^\star}$.
We then perform adversarial fine-tuning of $\mathcal{G}_{\theta}$ with an additional regularization term
\begin{equation}
    \mathcal{L}_{\mathrm{reg}}
    =
    \mathcal{L}_{\mathrm{HLF}}\big(\mathcal{G}_{\theta}(x^{\mathrm{LQ}}),\,\mathcal{G}_{\theta^\star}(x^{\mathrm{LQ}})\big),
    \label{eq:tpg_reg}
\end{equation}
where $\mathcal{L}_{\mathrm{HLF}}$ is the high-level feature (HLF) distance~\cite{kim2025exploiting}.
Intuitively, $\mathcal{L}_{\mathrm{HLF}}$ measures discrepancy between task-relevant representations of two images, so $\mathcal{L}_{\mathrm{reg}}$ discourages adversarial training from altering task-relevant details.
Concretely,
\begin{equation}
    \mathcal{L}_{\mathrm{HLF}}(x_{1},x_{2})
    =
    \frac{1}{2}
    \Big(
    \left\|\mathcal{H}^{\mathrm{feat}}_{\phi}(x_{1}) - \mathcal{H}^{\mathrm{feat}}_{\phi}(x_{2})\right\|_{1}
    +
    \left\|\bar{\mathcal{H}}^{\mathrm{feat}}(x_{1}) - \bar{\mathcal{H}}^{\mathrm{feat}}(x_{2})\right\|_{1}
    \Big),
    \label{eq:hlf}
\end{equation}
where $\mathcal{H}^{\mathrm{feat}}_{\phi}(\cdot)$ and $\bar{\mathcal{H}}^{\mathrm{feat}}(\cdot)$ are intermediate deep features extracted by $\mathcal{H}_\phi$ and a fixed HQ-pretrained task network $\bar{\mathcal{H}}$, respectively.
We term this training scheme task-preserving GAN training, illustrated in \cref{fig:tpgan}.

\subsection{Training Procedure}
\label{sec:method:training}

We follow an alternating optimization scheme~\cite{kim2024beyond, kim2025exploiting}, which updates the restoration network $\mathcal{G}_\theta$ (\cref{eq:loss_G}) and the task network $\mathcal{H}_\phi$ (\cref{eq:loss_F}) in turn for each training iteration.
When updating one network, we keep the other fixed (\ie, update $\mathcal{G}_\theta$ with $\mathcal{H}_\phi$ frozen, and vice versa).

\subsubsection{Update $\mathcal{G}_\theta$.}
Given $x^{\mathrm{LQ}}$, we restore $\hat{x}=\mathcal{G}_\theta(x^{\mathrm{LQ}})$ and optimize
\begin{equation}
    \min_{\theta}\;
    \mathcal{L}_{\mathrm{HLF}}(\hat{x}, x^{\mathrm{HQ}})
    +\alpha\,\mathcal{L}_{\mathrm{adv}}
    +\beta\,\mathcal{L}_{\mathrm{reg}},
    \label{eq:loss_G}
\end{equation}
where $\alpha$ and $\beta$ are weighting factors and $\mathcal{L}_{\mathrm{adv}}$ denotes the generator-side adversarial loss.
To obtain the reference model $\mathcal{G}_{\theta^\star}$ (without adversarial fine-tuning), we set $\alpha=\beta=0$ during its training.
When task-preserving GAN training is enabled, we additionally update a discriminator $\mathcal{D}_\psi$ using a standard adversarial objective~\cite{gans} to distinguish restored images $\hat{x}$ from HQ images.

Note that adopting NOLA not only stabilizes the training of the task network $\mathcal{H}_\phi$, but also benefits the training of $\mathcal{G}_\theta$.
Since $\mathcal{L}_{\mathrm{HLF}}$ is computed in the feature space of $\mathcal{H}_\phi$, deterministic forwarding stabilizes $\mathcal{H}^{\mathrm{feat}}_{\phi}$ and provides a more reliable optimization signal for training $\mathcal{G}_\theta$ to recover task-relevant details.

\vspace{-2mm}
\subsubsection{Update $\mathcal{H}_\phi$.}

To stabilize task-network training, we follow EDTR~\cite{kim2025exploiting} and construct each batch by mixing restored and HQ images in equal proportion, denoted $x^{\mathrm{mix}}$.
We then optimize
\begin{equation}
    \min_{\phi}\;
    \mathcal{L}_{\mathrm{task}}\big(\mathcal{H}_\phi(x^{\mathrm{mix}}), y\big)
    +\gamma\,\mathcal{L}_{\mathrm{FM}},
    \label{eq:loss_F}
\end{equation}
where $\mathcal{L}_{\mathrm{task}}$ is the standard task loss (\eg, cross-entropy for classification), $\gamma$ is a weighting factor, and $\mathcal{L}_{\mathrm{FM}} = \|\mathcal{H}^{\mathrm{feat}}_{\phi}(x^{\mathrm{mix}}) - \bar{\mathcal{H}}^{\mathrm{feat}}(x^{\mathrm{HQ}})\|_{1}$ is a feature-matching regularizer following EDTR~\cite{kim2025exploiting}.
We provide the detailed batch construction and full training procedure in Supplementary \cref{sec:additional_details}.

\section{Experiments}
\label{sec:experimnets}

\subsection{Main Benchmark Setup}

\subsubsection{Datasets.}
We use CUB-200-2011~\cite{WahCUB_200_2011} for image classification and PASCAL VOC 2012~\cite{everingham2010pascal} for semantic segmentation and object detection.
For each dataset, we use the official train/validation splits for training and validation, respectively.
To construct LQ inputs, we apply the degradation pipeline of Real-ESRGAN~\cite{rs_realesrgan} to the corresponding HQ images.

\vspace{-3mm}
\subsubsection{Compared Methods.}
\label{sssec:compared-methods}
We report results for three settings: (1) reference training without restoration, (2) IR baselines, and (3) TDIR methods.
As references, we train the task network using either HQ images (Upper Bound) or degraded LQ images without restoration (Lower Bound), and report the resulting task performance.
For IR baselines, we consider SwinIR~\cite{sr_swinir} and DiffBIR~\cite{lin2023diffbir}.
We train each IR model with its original objective, freeze it, and then train the task network using the restored images produced by the fixed IR model.
For TDIR baselines, including TDSR~\cite{sr_tdsr}, RSRSSN~\cite{zhao2018residual}, SR4IR~\cite{kim2024beyond}, and EDTR~\cite{kim2025exploiting}, we follow the original training protocols of the respective methods.
All methods are trained and evaluated on the datasets corresponding to each downstream task.

\vspace{-3mm}
\subsubsection{Implementation.}
For our IR model (NOLA-IR), we adopt SD v2.1~\cite{rombach2022high} as the backbone and apply LoRA~\cite{hu2022lora} adaptation with rank 64.
For the downstream task network ($\mathcal{H}_\phi$), we use ResNet~\cite{cls_resnet} for image classification, DeepLabV3~\cite{seg_deeplab} with a MobileNetV3-Large~\cite{howard2019searching} backbone for semantic segmentation, and Faster R-CNN~\cite{det_fasterrcnn} with a MobileNetV3-Large~\cite{howard2019searching} backbone for object detection.
Except for the reference settings, all task networks are initialized from the HQ-trained checkpoint, which we denote as HQ (Upper Bound), to ensure stable training and fair comparison.
We train all models for 10k iterations; further hyperparameters and training details are provided in the Supplementary \cref{sec:additional_details}.

\vspace{-3mm}
\subsubsection{Metrics.}
We report two types of metrics: downstream task performance and visual quality.
For task performance, we use top-1 accuracy (Acc, \%) for classification, mean intersection-over-union (mIoU, \%) for semantic segmentation, and mean average precision (mAP, \%) for object detection.
We report COCO~\cite{data_coco}-style mAP averaged over IoU thresholds from 0.5 to 0.95.
For visual quality, we use FID~\cite{heusel2017gans} and Q-Align~\cite{wu2023qalign} as perceptual metrics, together with PSNR as a distortion-based metric.
We compute FID between the distributions of restored images and the corresponding HQ images from the same dataset.

\subsection{Experimental Results on Main Benchmarks}

\begin{table*}[t!]
\caption{
    Comparison of high-level vision task performance and restored image quality under various image restoration methods.
    \textbf{Bold} and \underline{underlined} indicate the best and second-best results (excluding HQ upper bound), respectively.
}
\vspace{-2mm}
\small
\centering
\setlength\tabcolsep{1.0pt}
\def\arraystretch{1.1}
\label{table:main-table}
\resizebox{1.00\linewidth}{!}{
    \begin{tabular}{L{3.3cm}|C{1.3cm}|C{1.1cm}C{1.4cm}C{1.1cm}|C{1.3cm}|C{1.1cm}C{1.4cm}C{1.1cm}|C{1.3cm}|C{1.1cm}C{1.4cm}C{1.1cm}}
    \toprule
    \,~\multirow{3}{*}{Settings} & \multicolumn{4}{c|}{Image classification} & \multicolumn{4}{c|}{Semantic segmentation} & \multicolumn{4}{c}{Object detection} \\
    \cline{2-13}
    & Task & \multicolumn{3}{c|}{Visual quality} & Task & \multicolumn{3}{c|}{Visual quality} & Task & \multicolumn{3}{c}{Visual quality} \\
    & Acc$_\uparrow$ & FID$_\downarrow$ & Q-Align$_\uparrow$ & PSNR$_\uparrow$ & mIoU$_\uparrow$ & FID$_\downarrow$ & Q-Align$_\uparrow$ & PSNR$_\uparrow$ &  mAP$_\uparrow$ & FID$_\downarrow$ & Q-Align$_\uparrow$ & PSNR$_\uparrow$ \\
    \hline
    \;~HQ (Upper Bound) & \cellcolor{SpringGreen!\SpringGreenP}{82.5} & 0.00 & 3.69 & +$\inf$ & \cellcolor{SpringGreen!\SpringGreenP}{67.3} & 0.00 & 4.03 & +$\inf$ & \cellcolor{SpringGreen!\SpringGreenP}{37.1} & 0.00 & 3.95 & +$\inf$ \\
    \;~LQ (Lower Bound) & \cellcolor{SpringGreen!\SpringGreenP}{46.0} & 137.07 & 1.01 & 22.01 & \cellcolor{SpringGreen!\SpringGreenP}{34.3} & 125.16 & 1.08 & 21.17 & \cellcolor{SpringGreen!\SpringGreenP}{0.0*} & 136.95 & 1.02 & 20.07 \\
    \hline
    \;~SwinIR~\cite{sr_swinir} & \cellcolor{SpringGreen!\SpringGreenP}{58.2} & 66.97 & 1.93 & \textbf{23.96} & \cellcolor{SpringGreen!\SpringGreenP}{45.9} & 92.11 & 1.92 & \textbf{22.11} & \cellcolor{SpringGreen!\SpringGreenP}{19.4} & 86.73 & 1.71 & \textbf{20.92} \\
    \;~DiffBIR~\cite{lin2023diffbir} & \cellcolor{SpringGreen!\SpringGreenP}{60.6} & \underline{6.94} & \textbf{3.86} & 21.69 & \cellcolor{SpringGreen!\SpringGreenP}{50.0} & 49.41 & 3.52 & 20.68 & \cellcolor{SpringGreen!\SpringGreenP}{22.8} & 21.74 & 3.79 & 18.11 \\
\;~TDSR~\cite{sr_tdsr} & \cellcolor{SpringGreen!\SpringGreenP}{54.9} & 14.37 & 2.12 & \underline{23.22} & \cellcolor{SpringGreen!\SpringGreenP}{40.4} & 69.90 & 1.82 & \underline{21.95} & \cellcolor{SpringGreen!\SpringGreenP}{13.1} & 54.95 & 1.73 & \underline{20.74} \\
    \;~RSRSSN~\cite{zhao2018residual} & \cellcolor{SpringGreen!\SpringGreenP}{61.3} & 19.25 & 2.49 & 20.69 & \cellcolor{SpringGreen!\SpringGreenP}{46.1} & 72.69 & 1.92 & 18.79 & \cellcolor{SpringGreen!\SpringGreenP}{19.9} & 56.57 & 1.76 & 18.49 \\
    \;~SR4IR~\cite{kim2024beyond} & \cellcolor{SpringGreen!\SpringGreenP}{63.8} & 8.05 & 3.10 & 22.71 & \cellcolor{SpringGreen!\SpringGreenP}{46.5} & 53.89 & 2.99 & 21.44 & \cellcolor{SpringGreen!\SpringGreenP}{20.0} & 41.92 & 2.93 & 20.38 \\
    \;~EDTR~\cite{kim2025exploiting} & \cellcolor{SpringGreen!\SpringGreenP}{\underline{67.2}} & 7.35 & 3.70 & 21.75 & \cellcolor{SpringGreen!\SpringGreenP}{\underline{58.7}} & \underline{35.52} & \underline{4.08} & 20.23 & \cellcolor{SpringGreen!\SpringGreenP}{\underline{30.2}} & \underline{16.96} & \textbf{3.95} & 19.14 \\
    \;~\textbf{NOLA-IR (Ours)} & \cellcolor{SpringGreen!\SpringGreenP}{\textbf{69.3}} & \textbf{4.07} & \underline{3.83} & 22.49 & \cellcolor{SpringGreen!\SpringGreenP}{\textbf{62.9}} & \textbf{29.59} & \textbf{4.11} & 20.66 & \cellcolor{SpringGreen!\SpringGreenP}{\textbf{32.0}} & \textbf{12.44} & \underline{3.87} & 19.75 \\
    \bottomrule
    \end{tabular}
}
\caption*{\hfill\scriptsize * Training did not converge.}
\vspace{-8mm}
\end{table*}

\begin{figure*}[t!]
    \centering
    \captionsetup[subfigure]{labelfont=scriptsize, textfont=scriptsize}
    \renewcommand{\wp}{0.240}
    \renewcommand{\splitT}{0.250}

    \begin{tikzpicture}[remember picture]
        \node[inner sep=0pt] (grid) {%
            \begin{minipage}{\textwidth}
                \centering
                \subfloat[LQ~(Lower Bound)]{\includegraphics[width=\wp\linewidth]{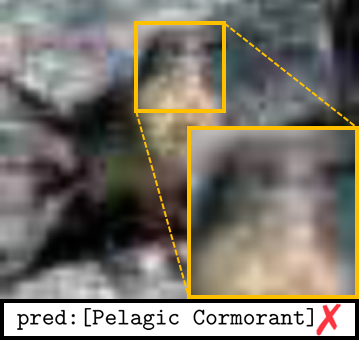}}
                \vspace{1mm}
                \hfill
                \subfloat[SwinIR~\cite{sr_swinir}]{\includegraphics[width=\wp\linewidth]{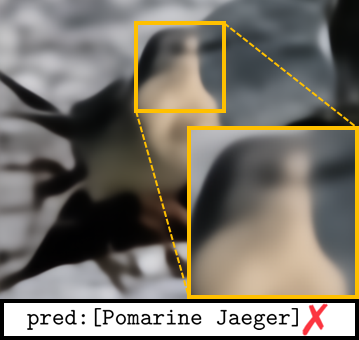}}
                \hfill
                \subfloat[DiffBIR~\cite{lin2023diffbir}]{\includegraphics[width=\wp\linewidth]{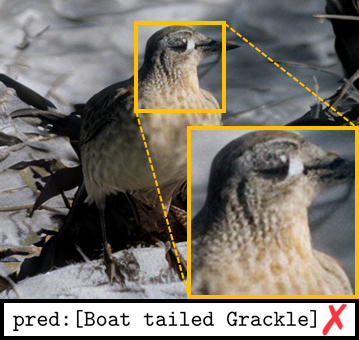}}
                \hfill
                \subfloat[SR4IR~\cite{kim2024beyond}]{\includegraphics[width=\wp\linewidth]{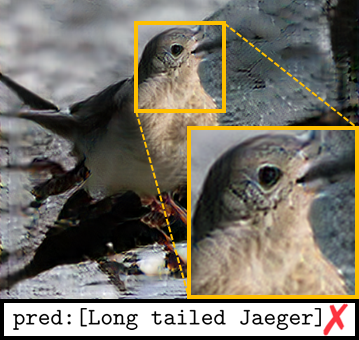}}

                \par\vspace{2mm}

                \subfloat[HQ~(Ground-truth)]{\includegraphics[width=\wp\linewidth]{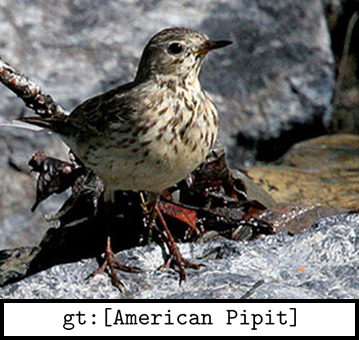}}
                \vspace{1mm}
                \hfill
                \subfloat[EDTR~\cite{kim2025exploiting}]{\includegraphics[width=\wp\linewidth]{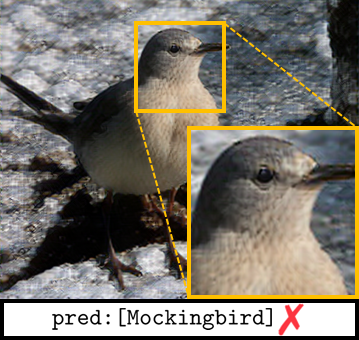}}
                \hfill
                \subfloat[\textbf{NOLA-IR}]{\includegraphics[width=\wp\linewidth]{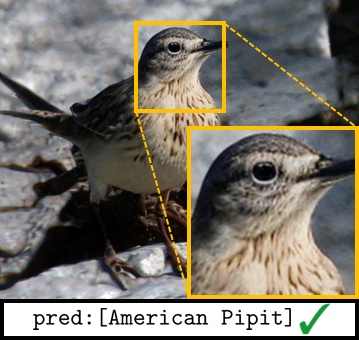}}
                \hfill
                \subfloat[HQ~(Upper Bound)]{\includegraphics[width=\wp\linewidth]{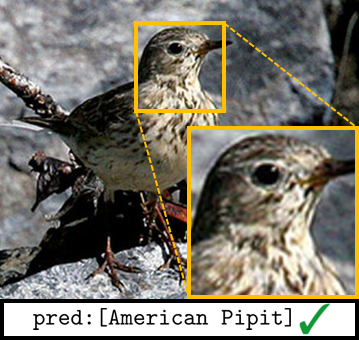}}
            \end{minipage}%
        };

        \begin{scope}[overlay]
            \draw[dashed, line width=0.8pt]
              ($(grid.north west)!\splitT!(grid.north east)$) -- ($(grid.south west)!\splitT!(grid.south east)$);
        \end{scope}
    \end{tikzpicture}
    
    \addtocounter{subfigure}{-8}
    
    \vspace{-2.5mm}
    \caption{Qualitative results on CUB-200-2011 with downstream classification outcomes. HQ with Ground-truth label denotes the original image with the ground-truth label, while Upper Bound label reports the prediction of a task network trained on HQ images.}
    \label{fig:vis-cls}
\end{figure*}

\begin{figure*}[t!]
    \centering
    \captionsetup[subfigure]{labelfont=scriptsize, textfont=scriptsize}
    \renewcommand{\wp}{0.240}
    \renewcommand{\splitT}{0.250}

    \begin{tikzpicture}[remember picture]
        \node[inner sep=0pt] (grid) {%
            \begin{minipage}{\textwidth}
                \centering
                \subfloat[LQ (Lower Bound)]{\includegraphics[width=\wp\linewidth]{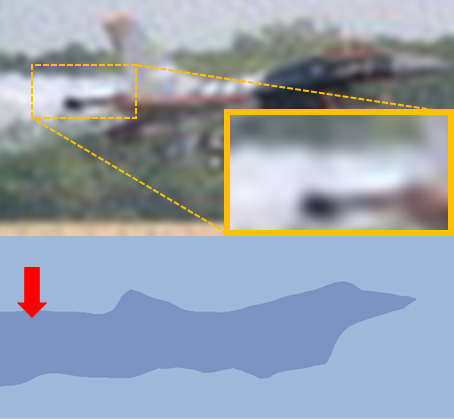}}
                \vspace{1mm}
                \hfill
                \subfloat[SwinIR~\cite{sr_swinir}]{\includegraphics[width=\wp\linewidth]{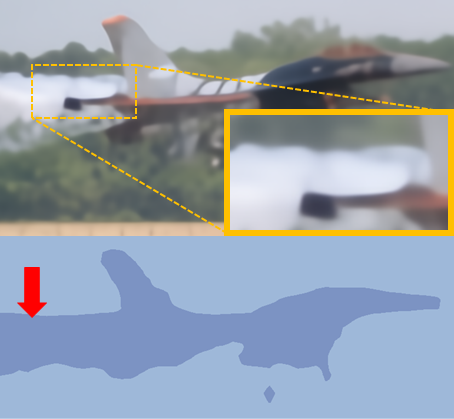}}
                \hfill
                \subfloat[DiffBIR~\cite{lin2023diffbir}]{\includegraphics[width=\wp\linewidth]{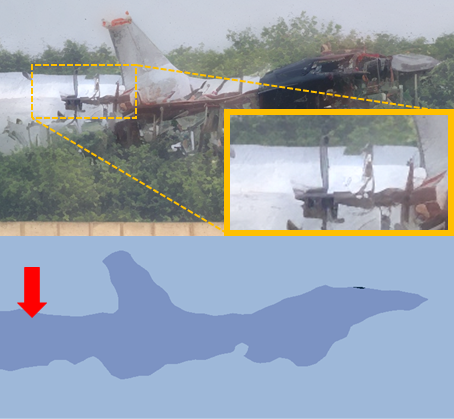}}
                \hfill
                \subfloat[SR4IR~\cite{kim2024beyond}]{\includegraphics[width=\wp\linewidth]{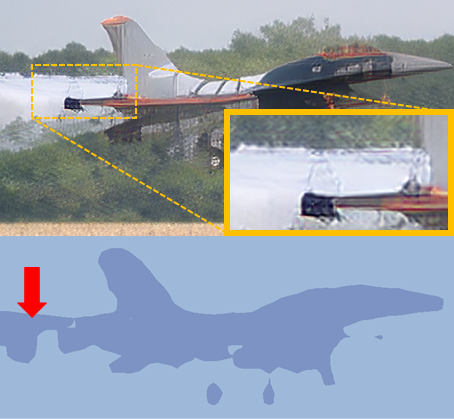}}


                \subfloat[HQ~(Ground-truth)]{\includegraphics[width=\wp\linewidth]{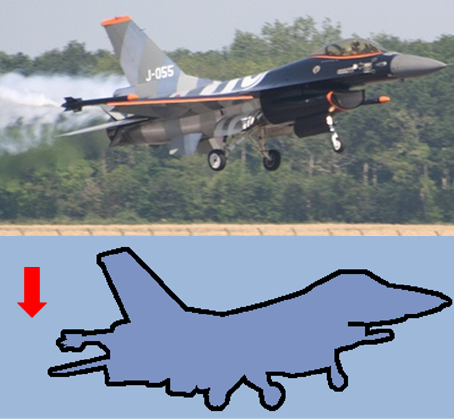}}
                \vspace{1mm}
                \hfill
                \subfloat[EDTR~\cite{kim2025exploiting}]{\includegraphics[width=\wp\linewidth]{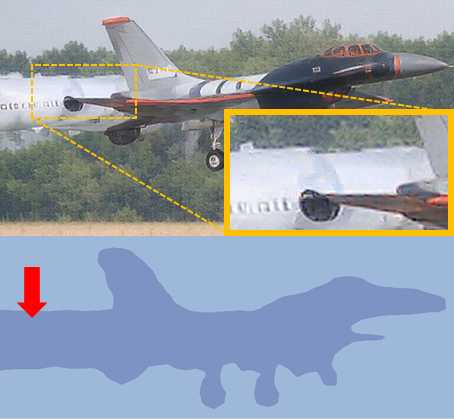}}
                \hfill
                \subfloat[\textbf{NOLA-IR}]{\includegraphics[width=\wp\linewidth]{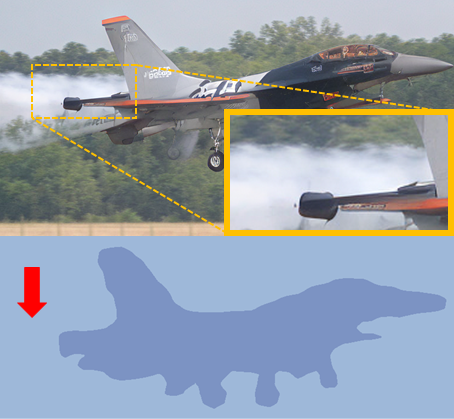}}
                \hfill
                \subfloat[HQ~(Upper Bound)]{\includegraphics[width=\wp\linewidth]{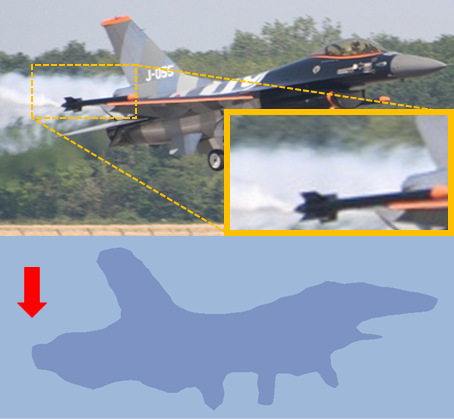}}
            \end{minipage}%
        };

        \begin{scope}[overlay]
            \draw[dashed, line width=0.8pt]
              ($(grid.north west)!\splitT!(grid.north east)$) -- ($(grid.south west)!\splitT!(grid.south east)$);
        \end{scope}
    \end{tikzpicture}
    
    \addtocounter{subfigure}{-8}
    
    \vspace{-2.5mm}
    \caption{Qualitative results on PASCAL VOC 2012 with downstream segmentation outcomes. Notation follows \cref{fig:vis-cls}. In the ground-truth segmentation map (e), black denotes the ``Don't Care'' region.}
    \label{fig:vis-seg}
\vspace{-5mm}
\end{figure*}

\cref{table:main-table} reports quantitative comparisons across three downstream tasks (classification, segmentation, detection), and \cref{fig:vis-cls,fig:vis-seg,fig:vis-det} provide qualitative results with the corresponding task predictions.
Overall, excluding the HQ upper-bound reference, our method achieves the best task performance across all three tasks, while also maintaining strong perceptual quality as reflected by FID and Q-Align.
We note that distortion-oriented methods~\cite{sr_swinir} may achieve higher PSNR, but PSNR can be less indicative of downstream task performance~\cite{kim2024beyond, pei2018does, dai2016image}.
Beyond task performance, NOLA-IR is also efficient: with one-step forwarding and parameter-efficient LoRA, it reduces inference cost by over an order of magnitude compared to multi-step diffusion restoration (\eg, 0.22\,s vs.\ 4.8\,s for DiffBIR), as detailed in Supplementary \cref{sec:computational_cost}.

\vspace{-3mm}
\subsubsection{Classification.}
Our NOLA-IR achieves the best classification accuracy among compared restoration methods, improving the previous state-of-the-art TDIR method EDTR~\cite{kim2025exploiting} from 67.2 to 69.3 (+2.1 Acc).
This gain supports our main claim that using a deterministic one-step forward pass helps the task network learn more stable decision boundaries.
In addition, incorporating task-preserving GAN training improves perceptual quality, reflected by the best FID and the second-best Q-Align.
As shown in \cref{fig:vis-cls}, NOLA-IR restores fine-grained structures (\eg, facial details and feather patterns) that are crucial for bird species classification, leading to correct predictions in cases where other methods fail.

\subsubsection{Segmentation.}
NOLA-IR achieves the best mIoU among the compared restoration methods, improving EDTR from 58.7 to 62.9 (+4.2 mIoU), while also attaining the best FID and Q-Align.
As shown in \cref{fig:vis-seg}, NOLA-IR better recovers the fine structures and boundary between the airplane body (\eg, rear engine area) and the trailing contrails, leading to a clean separation in the segmentation output.
In contrast, compared methods fail to restore these details, causing the airplane and contrails to be merged into a single region.

\begin{figure*}[t!]
    \centering
    \captionsetup[subfigure]{labelfont=scriptsize, textfont=scriptsize}
    \renewcommand{\wp}{0.240}
    \renewcommand{\splitT}{0.250}

    \begin{tikzpicture}[remember picture]
        \node[inner sep=0pt] (grid) {%
            \begin{minipage}{\textwidth}
                \centering
                \subfloat[LQ (Lower Bound)]{\includegraphics[width=\wp\linewidth]{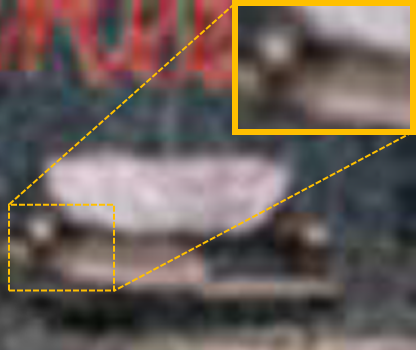}}
                \vspace{1mm}
                \hfill
                \subfloat[SwinIR~\cite{sr_swinir}]{\includegraphics[width=\wp\linewidth]{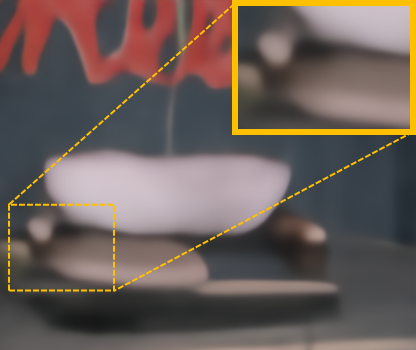}}
                \hfill
                \subfloat[DiffBIR~\cite{lin2023diffbir}]{\includegraphics[width=\wp\linewidth]{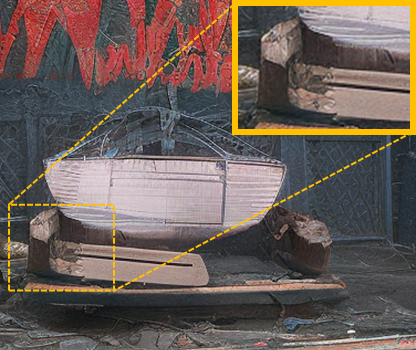}}
                \hfill
                \subfloat[SR4IR~\cite{kim2024beyond}]{\includegraphics[width=\wp\linewidth]{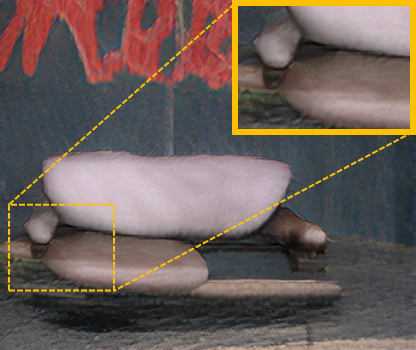}}

                \par\vspace{2mm}

                \subfloat[HQ~(Ground-truth)]{\includegraphics[width=\wp\linewidth]{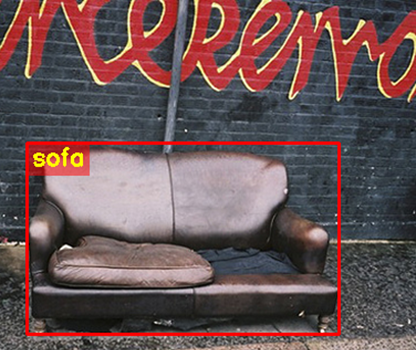}}
                \vspace{1mm}
                \hfill
                \subfloat[EDTR~\cite{kim2025exploiting}]{\includegraphics[width=\wp\linewidth]{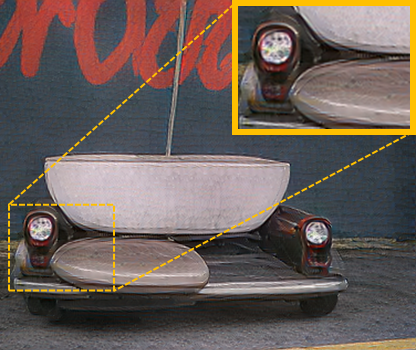}}
                \hfill
                \subfloat[\textbf{NOLA-IR}]{\includegraphics[width=\wp\linewidth]{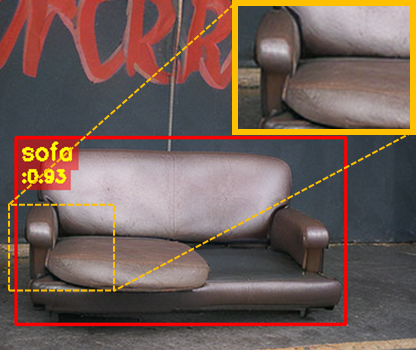}}
                \hfill
                \subfloat[HQ~(Upper Bound)]{\includegraphics[width=\wp\linewidth]{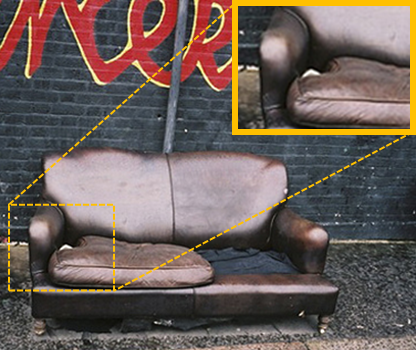}}
            \end{minipage}%
        };

        \begin{scope}[overlay]
            \draw[dashed, line width=0.8pt]
              ($(grid.north west)!\splitT!(grid.north east)$) -- ($(grid.south west)!\splitT!(grid.south east)$);
        \end{scope}
    \end{tikzpicture}

    \addtocounter{subfigure}{-8}
    \vspace{-2mm}
    \caption{Qualitative results on PASCAL VOC 2012 with downstream detection outcomes. Notation follows \cref{fig:vis-cls}.}
    \label{fig:vis-det}
    \vspace{-5mm}
\end{figure*}

\vspace{-2mm}
\subsubsection{Detection.}
NOLA-IR yields the best mAP among the compared restoration methods, improving EDTR from 30.2 to 32.0 (+1.8 mAP).
In terms of visual quality, NOLA-IR again achieves the best FID and a strong Q-Align score.
As shown in \cref{fig:vis-det}, NOLA-IR restores fine details of the sofa and enables correct detection, whereas other methods fail to detect it.
Interestingly, even the HQ-trained (Upper Bound) model fails in this example, suggesting that task-driven restoration can produce task-friendlier cues than the original HQ input.
Additional qualitative results for all tasks are provided in Supplementary \cref{sec:visualization_main}.

\subsection{Analysis}

\subsubsection{Effect of Noise-Free Forward Pass and Adapter Choice.}
\label{ssec:abl-nola}

We use a noise-free one-step forward pass (\cref{eq:one-step}) when leveraging a pretrained diffusion prior, which enables deterministic restoration.
\cref{table:nola-ablation} summarizes its effect under different adaptation modules.
To isolate the effect of the noise-free forward pass, we disable task-preserving GAN training in this analysis.

We start from a baseline that follows the one-step EDTR setting~\cite{kim2025exploiting}, using a ControlNet-style adapter (IRControlNet~\cite{lin2023diffbir}) with mildly noisy latent initialization.
Applying the noise-free forward pass to this ControlNet baseline does not improve performance and even slightly degrades it (up to $-2.5$ mIoU on segmentation).
We then replace only the adaptation module (ControlNet $\rightarrow$ LoRA), keeping all other settings identical (\eg, backbone, training protocol).
Under this change, the same noise-free forward pass yields consistent gains across all tasks.

To understand why the two adapters behave differently, we further report the Task-Feature Distance (TFD),
\begin{equation}
d_{\mathrm{task}}(\hat{x}, x^{\mathrm{HQ}}) = \big\|\bar{\mathcal{H}}^{\mathrm{feat}}(\hat{x}) - \bar{\mathcal{H}}^{\mathrm{feat}}(x^{\mathrm{HQ}})\big\|_{1},
\label{eq:tfd}
\end{equation}
which measures how closely the restored image $\hat{x}$ matches the HQ image in the task-feature space of a fixed HQ-pretrained task network $\bar{\mathcal{H}}$.
A lower TFD indicates that more task-relevant details are recovered.
As shown in \cref{table:nola-ablation}, noise-free forwarding with LoRA consistently reduces TFD across all tasks, aligning with its improved task performance.
In contrast, with ControlNet the same forwarding fails to reduce TFD, consistent with its lack of gains.
These results show that the noise-free forward pass improves the recovery of task-relevant details---and thus downstream task performance---only when paired with a LoRA adapter.

\begin{table}[t!]
\caption{
    Effect of noise-free one-step forward pass under different adaptation modules. All settings share the same SD-based backbone and training protocol. Only the adaptation module and the presence of the random noise term differ.
}
\vspace{-2mm}
\small
\centering
\setlength\tabcolsep{1.0pt}
\def\arraystretch{1.1}
\label{table:nola-ablation}
\resizebox{0.95\linewidth}{!}{
    \begin{tabular}{L{5.6cm}|C{1.7cm}C{1.4cm}|C{1.7cm}C{1.4cm}|C{1.7cm}C{1.4cm}}
    \toprule
    \;~\multirow{2}{*}{Methods} & \multicolumn{2}{c|}{Classification} &  \multicolumn{2}{c|}{Segmentation} & \multicolumn{2}{c}{Detection} \\
    \cline{2-7}
    & Acc$_\uparrow$ & TFD$_\downarrow$ & mIoU$_\uparrow$ & TFD$_\downarrow$ & mAP$_\uparrow$ & TFD$_\downarrow$ \\
    \hline
    \;~Baseline (ControlNet) & 67.1 & 0.544 & 56.8 & 0.471 & 28.4 & 0.699 \\
    \;~+Noise-Free Forward Pass & 66.7 \textcolor{red4mark}{(-0.4)} & 0.544 & 54.3 \textcolor{red4mark}{(-2.5)} & 0.474 & 28.2 \textcolor{red4mark}{(-0.2)} & 0.699 \\
    \hline
    \;~Baseline (LoRA) & 66.9 & 0.532 & 60.1 & 0.461 & 30.0 & 0.688 \\
    \;~+Noise-Free Forward Pass (\textbf{NOLA}) & 69.2 \textcolor{green4mark}{(+2.3)} & 0.519 & 62.6 \textcolor{green4mark}{(+2.5)} & 0.444 & 32.1 \textcolor{green4mark}{(+2.1)} & 0.666 \\
    \bottomrule
    \end{tabular}
}
\end{table}

\begin{table}[t!]
\caption{Sensitivity to input perturbations under the noise-free setting (segmentation).}
\vspace{-2mm}
\small
\centering
\setlength\tabcolsep{3pt}
\def\arraystretch{1.1}
\label{table:sensitivity}
\resizebox{0.95\linewidth}{!}{
    \begin{tabular}{L{8.5cm}|C{2.2cm}|C{1.6cm}}
    \toprule
    \;~Methods & Sensitivity $\downarrow$ & mIoU $\uparrow$ \\
    \hline
    \;~Baseline~(ControlNet) + Noise-Free Forward Pass & 0.4594 & 54.3 \\
    \;~Baseline~(LoRA) + Noise-Free Forward Pass (\textbf{NOLA}) & \textbf{0.3742} & \textbf{62.6} \\
    \bottomrule
    \end{tabular}
}
\vspace{-3mm}
\end{table}

\vspace{-2mm}
\paragraph{Discussion.}
Why does the same noise-free forward pass behave differently across adapters?
We attribute this to an input-domain shift: removing the noise term drives the denoiser input away from the noisy-latent distribution seen during SD pretraining, and adapters differ in their capacity to cope with this shift.
ControlNet steers a frozen denoiser through additive conditioning, which is limited in compensating for this mismatch under the noise-free latent distribution.
In contrast, LoRA directly updates the denoiser weights in a parameter-efficient manner, enabling more effective adaptation under the noise-free regime.

To further probe this, we measure each adapter's sensitivity to input perturbations under the noise-free setting.
Specifically, we perturb the input with Gaussian noise and compute the relative change of intermediate restoration features, $\|\mathcal{G}^{\mathrm{feat}}(x) - \mathcal{G}^{\mathrm{feat}}(x^{\mathrm{pert}})\|_2^2 / \|\mathcal{G}^{\mathrm{feat}}(x)\|_2^2$.
As shown in \cref{table:sensitivity}, LoRA is substantially less sensitive to input perturbations than ControlNet.
This greater input stability is consistent with our input-domain-shift explanation: an adapter that responds more stably to input variations is better positioned to absorb the distribution shift introduced by noise-free forwarding, in line with LoRA's lower TFD and higher task performance.

We provide additional comparisons in Supplementary \cref{sec:additional_ablations}, including (i) parameter-matched ControlNet vs.\ LoRA and (ii) using the original ControlNet instead of IRControlNet, and observe consistent trends.

\begin{table}[t!]
\caption{
    Ablation of task-preserving GAN training for NOLA-IR on object detection.
}
\vspace{-2mm}
\small
\centering
\setlength\tabcolsep{1.0pt}
\def\arraystretch{1.1}
\label{table:task-preserving-gan-training}
\resizebox{0.95\linewidth}{!}{
    \begin{tabular}{L{7.0cm}|C{1.5cm}|C{1.5cm}|C{1.5cm}|C{1.7cm}|C{1.7cm}|C{1.7cm}}
    \toprule
    \;~\multirow{2}{*}{Methods} & \multicolumn{2}{c|}{Task} & \multicolumn{4}{c}{Visual quality} \\
    \cline{2-7}
    & mAP$_\uparrow$ & mAP$_{0.5}$$_\uparrow$ & FID$_\downarrow$ & Q-Align$_\uparrow$ & MANIQA$_\uparrow$ & MUSIQ$_\uparrow$ \\
    \hline
    \;~Baseline & \cellcolor{SpringGreen!\SpringGreenP}{32.1} & \cellcolor{SpringGreen!\SpringGreenP}{54.2} & 17.19 & 3.42 & 0.438 & 68.49 \\
    \;~+GAN Training & \cellcolor{SpringGreen!\SpringGreenP}{31.2} & \cellcolor{SpringGreen!\SpringGreenP}{53.2} & 12.32 & 3.89 & 0.483 & 69.25 \\
    \;~+Task-Preserving GAN Training (\textbf{NOLA-IR}) & \cellcolor{SpringGreen!\SpringGreenP}{32.0} & \cellcolor{SpringGreen!\SpringGreenP}{54.4} & 12.44 & 3.87 & 0.484 & 69.28 \\
    \bottomrule
    \end{tabular}
}
\vspace{-1mm}
\end{table}

\begin{table}[t!]
\centering
\begin{minipage}[t]{0.493\linewidth}
    \centering
    \caption{Comparison with multi-sample aggregation on segmentation. NFE denotes the number of function evaluations; we use the one-step EDTR model.}
    \vspace{-2mm}
    \small
    \setlength\tabcolsep{3pt}\def\arraystretch{1.1}
    \label{table:avg4}
    \resizebox{\linewidth}{!}{
        \begin{tabular}{L{4.8cm}|C{1.3cm}|C{1.3cm}}
        \toprule
        Methods & mIoU$_\uparrow$ & NFE$_\downarrow$ \\
        \hline
        EDTR~\cite{kim2025exploiting} & 57.3 & 1 \\
        EDTR~\cite{kim2025exploiting} (Avg over 4 samples) & 58.5 & 4 \\
        \textbf{NOLA-IR (Ours)} & \textbf{62.9} & 1 \\
        \bottomrule
        \end{tabular}
    }
\end{minipage}
\hfill
\begin{minipage}[t]{0.467\linewidth}
    \centering
    \caption{Comparison with one-step diffusion IR methods on segmentation.}
    \vspace{-2mm}
    \small
    \setlength\tabcolsep{3pt}\def\arraystretch{1.1}
    \label{table:osediff}
    \resizebox{\linewidth}{!}{
        \begin{tabular}{L{5.0cm}|C{1.6cm}}
        \toprule
        Methods & mIoU$_\uparrow$ \\
        \hline
        OSEDiff~\cite{wu2024one} & 54.8 \\
        OSEDiff~\cite{wu2024one} (Task-supervised) & 60.3 \\
        InvSR~\cite{yue2025arbitrary} & 34.9 \\
        TSD-SR~\cite{dong2025tsd} & 41.5 \\
        \textbf{NOLA-IR (Ours)} & \textbf{62.9} \\
        \bottomrule
        \end{tabular}
    }
\end{minipage}
\vspace{-3mm}
\end{table}

\vspace{-4mm}
\subsubsection{Effect of Task-Preserving GAN Training.}
\label{ssec:abl-tpgan}

\cref{table:task-preserving-gan-training} reports the effectiveness of our task-preserving GAN training (\cref{sec:method:tp-gan}) on object detection.
In addition to COCO-style mAP, we also report mAP$_{0.5}$ (AP at an IoU threshold of 0.5) for task performance.
For visual quality, we include MANIQA~\cite{yang2022maniqa} and MUSIQ~\cite{ke2021musiq} in addition to FID and Q-Align for a more comprehensive evaluation.

The baseline~(first row) corresponds to NOLA-IR trained without adversarial loss.
Adding a standard adversarial objective (second row) substantially improves perceptual metrics (\eg, lower FID and higher Q-Align/MANIQA/MUSIQ), but it also degrades detection accuracy, indicating a tension between adversarial objectives and task-driven optimization.
In contrast, adopting task-preserving GAN training (third row) retains most of the perceptual gains while preserving task performance, demonstrating that the proposed regularization effectively balances the two objectives.

\vspace{-4mm}
\subsubsection{Comparison with Multi-Sample Aggregation.}
\label{ssec:abl-avg}

A natural alternative to eliminating stochasticity is to average multiple stochastic restorations.
We compare NOLA-IR against aggregating the predictions of four stochastic one-step EDTR~\cite{kim2025exploiting} runs (\cref{table:avg4}).
Averaging improves mIoU over a single stochastic run, confirming that stochasticity indeed harms task prediction.
However, it still underperforms our deterministic NOLA-IR (58.5 vs.\ 62.9 mIoU) while requiring four times the computation (NFE~4 vs.\ 1).
Thus, multi-sample aggregation only mitigates stochasticity, matching neither the effectiveness nor the efficiency of our deterministic forwarding.

\vspace{-4mm}
\subsubsection{Comparison with One-Step Diffusion Methods.}
\label{ssec:abl-osediff}

We compare NOLA-IR against recent one-step diffusion IR methods---OSEDiff~\cite{wu2024one}, InvSR~\cite{yue2025arbitrary}, and TSD-SR~\cite{dong2025tsd}---under our segmentation protocol (\cref{table:osediff}).
We further train a task-supervised variant of OSEDiff, which adds a segmentation task loss to its original restoration objective following TDSR~\cite{sr_tdsr}.
NOLA-IR achieves the best mIoU; notably, even task-supervised OSEDiff falls short, confirming that our gains stem from the task-driven framework rather than one-step diffusion alone.

\begin{figure*}[t!]
    \centering
    \captionsetup[subfigure]{labelfont=scriptsize, textfont=scriptsize}
    \renewcommand{\wp}{0.33}
    \renewcommand{\splitT}{0.250}
    \resizebox{0.8\linewidth}{!}{
        \begin{minipage}{\textwidth}
            \centering
            \vspace{1mm}
            \addtocounter{subfigure}{-3}
            \subfloat{\includegraphics[width=0.19\linewidth]{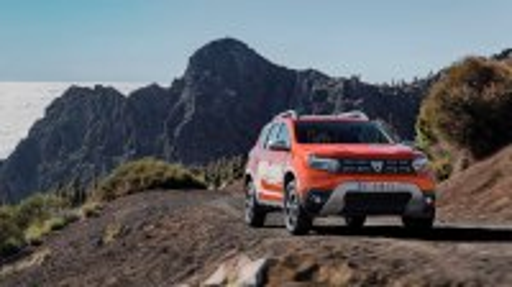}}
            \hfill
            \subfloat{\includegraphics[width=0.39\linewidth]{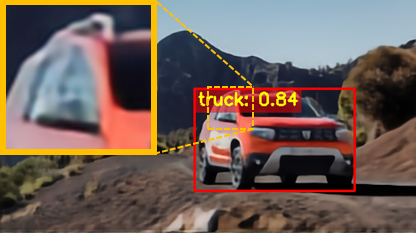}}
            \hfill
            \subfloat{\includegraphics[width=0.39\linewidth]{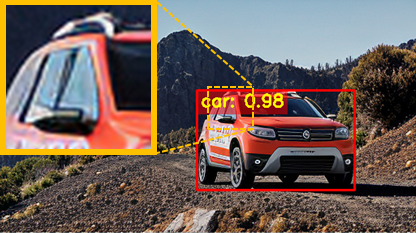}}
            \\
            \vspace{1mm}
            \addtocounter{subfigure}{-3}
            \subfloat[LQ]{\includegraphics[width=0.19\linewidth]{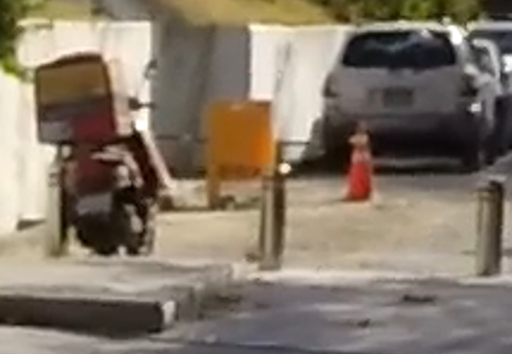}}
            \hfill
            \subfloat[SwinIR~\cite{sr_swinir}]{\includegraphics[width=0.39\linewidth]{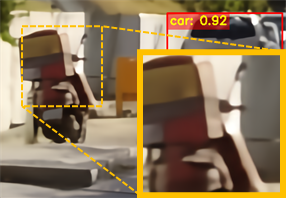}}
            \hfill
            \subfloat[\textbf{NOLA-IR}]{\includegraphics[width=0.39\linewidth]{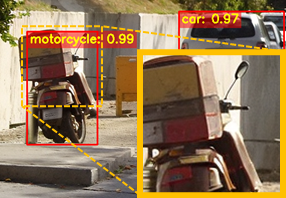}}
        \end{minipage}%
        \addtocounter{subfigure}{-3}
    }
    \vspace{-1mm}
    \caption{Restoration results and detection outcomes on real-world LQ images. Each row shows (a) a real-world LQ input and the corresponding detector outputs after restoration using (b) SwinIR~\cite{sr_swinir} or (c) NOLA-IR. The two examples are from RealLR200~\cite{wu2024seesr} and a cropped patch from GOPRO~\cite{db_deepdeblur}, respectively.
    }
    \label{fig:vis-real-world-det}
\vspace{-3mm}
\end{figure*}

\vspace{-3mm}
\subsection{Generalization Beyond the Main Benchmarks}

\subsubsection{Real-World Application on Detection.}
\label{ssec:main-paper-real-world-det}
To evaluate generalization to real-world LQ inputs with unknown degradations, we apply our method to object detection on real-world LQ images~\cite{wu2024seesr, db_deepdeblur}.
Instead of reusing models from the main benchmark setup, we retrain NOLA-IR on the larger-scale MS COCO 2017 dataset~\cite{data_coco} with a stronger Faster R-CNN detector using a ResNet-50 backbone.
During training, we synthesize LQ inputs using the same Real-ESRGAN degradation pipeline as in the main benchmark setup, and train for 200k iterations.

\cref{fig:vis-real-world-det} shows qualitative results on real-world LQ images, comparing against SwinIR~\cite{sr_swinir} retrained on COCO under the same protocol (\cref{sssec:compared-methods}).
Across diverse degradations, NOLA-IR better recovers task-relevant details that SwinIR misses---\eg, fine car details and the motorcycle side mirror---enabling correct detections where SwinIR fails.
Additional real-world results are in Supplementary \cref{sec:visualization_real_detection}, and quantitative COCO-val results in Supplementary \cref{sec:quantitative_coco}, where NOLA-IR substantially improves detection over SwinIR (\eg, mAP 16.4\,$\rightarrow$\,24.5).

\vspace{-3mm}
\subsubsection{Extension to OCR.}
\label{ssec:main-paper-ocr}
To test generalization beyond standard high-level vision tasks, we apply NOLA-IR to OCR.
We train on MJSynth~\cite{jaderberg2014synthetic} with a CRNN~\cite{shi2016end}-based OCR network~\cite{baek2019wrong}, using the same Real-ESRGAN degradation pipeline as before (training details in Supplementary \cref{sssec:additional_details_training_ocr}).
We report case-insensitive word accuracy (exact match) and character accuracy (position-wise match ratio normalized by the longer string length), together with FID and PSNR.

\begin{figure*}[t!]
    \centering
    \captionsetup[subfigure]{labelfont=scriptsize, textfont=scriptsize}
    \renewcommand{\wp}{0.24}
    \renewcommand{\splitT}{0.250}
    \resizebox{1.0\linewidth}{!}{
        \begin{minipage}{\textwidth}
            \centering
            \addtocounter{subfigure}{-4}
            \subfloat{\includegraphics[width=\wp\linewidth]{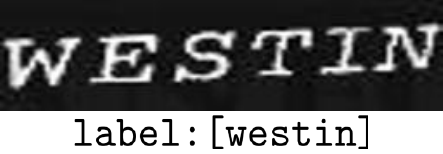}}
            \hfill
            \subfloat{\includegraphics[width=\wp\linewidth]{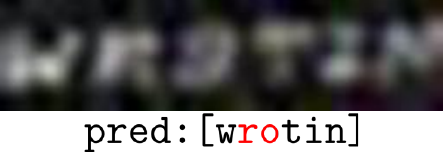}}
            \hfill
            \subfloat{\includegraphics[width=\wp\linewidth]{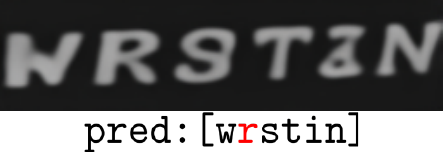}}
            \hfill
            \subfloat{\includegraphics[width=\wp\linewidth]{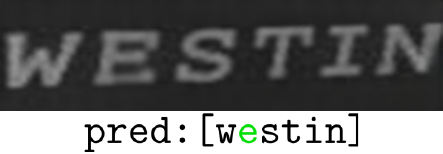}}
            \\
            \vspace{1mm}
            \addtocounter{subfigure}{-4}
            \subfloat{\includegraphics[width=\wp\linewidth]{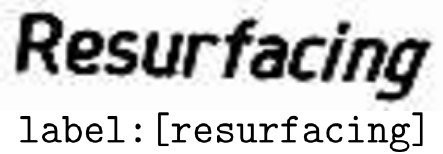}}
            \hfill
            \subfloat{\includegraphics[width=\wp\linewidth]{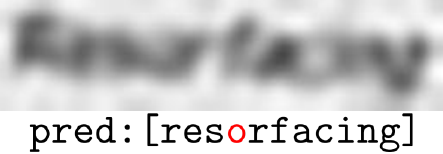}}
            \hfill
            \subfloat{\includegraphics[width=\wp\linewidth]{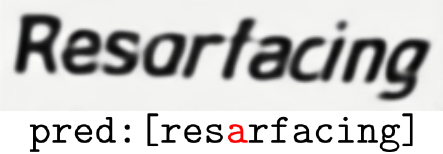}}
            \hfill
            \subfloat{\includegraphics[width=\wp\linewidth]{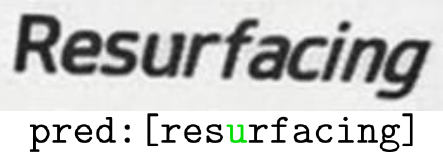}}
            \\
            \vspace{1mm}
            \addtocounter{subfigure}{-4}
            \subfloat[HQ (Ground-truth)]{\includegraphics[width=\wp\linewidth]{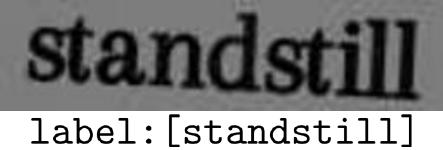}}
            \hfill
            \subfloat[LQ (Lower Bound)]{\includegraphics[width=\wp\linewidth]{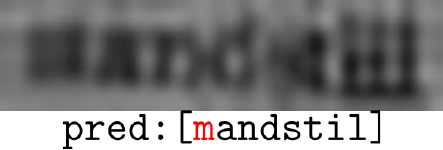}}
            \hfill
            \subfloat[SwinIR~\cite{sr_swinir}]{\includegraphics[width=\wp\linewidth]{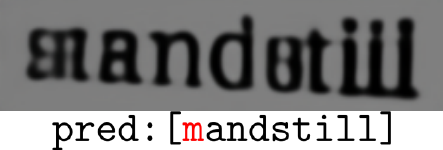}}
            \hfill
            \subfloat[\textbf{NOLA-IR}]{\includegraphics[width=\wp\linewidth]{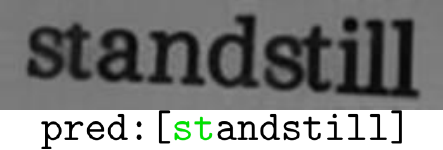}}
        \end{minipage}%
        \addtocounter{subfigure}{-8}
    }

    \vspace{-2mm}
    \caption{Qualitative OCR results on MJSynth~\cite{jaderberg2014synthetic}. Each row shows (a) an HQ word image with its ground-truth text, (b) its degraded LQ counterpart, and restorations by (c) SwinIR~\cite{sr_swinir} and (d) NOLA-IR, together with the corresponding OCR predictions.}
    \label{fig:vis-ocr}
\end{figure*}

\cref{table:ocr-table} reports quantitative results.
OCR performance drops sharply on degraded inputs (\eg, word accuracy from 95.7 to 20.6), confirming that OCR is highly sensitive to image quality.
NOLA-IR improves both word and character accuracy over the compared methods while also achieving the best FID, demonstrating that our approach generalizes to OCR.
As shown in \cref{fig:vis-ocr}, NOLA-IR restores more legible character structures than others---for example, correctly recovering the word ``westin'', which is misrecognized under the LQ and SwinIR settings.
Additional qualitative examples are provided in Supplementary \cref{sec:visualization_ocr}.

\begin{table}[t!]
\caption{
    OCR performance and restoration quality on MJSynth. \textbf{Bold} indicates the best result excluding the HQ upper bound.
}
\vspace{-3mm}
\small
\centering
\setlength\tabcolsep{1.0pt}
\def\arraystretch{1.1}
\label{table:ocr-table}
\resizebox{0.8\linewidth}{!}{
    \begin{tabular}{L{4.0cm}|C{2.0cm}C{2.5cm}|C{1.8cm}C{1.8cm}}
    \toprule
    \,~\multirow{2}{*}{IR methods} & \multicolumn{2}{c|}{Task} & \multicolumn{2}{c}{Visual quality} \\
    & Word Acc$_\uparrow$ & Character Acc$_\uparrow$ & FID$_\downarrow$ & PSNR$_\uparrow$ \\
    \hline
    \;~HQ~(Upper Bound) & \cellcolor{SpringGreen!\SpringGreenP}{95.7} & \cellcolor{SpringGreen!\SpringGreenP}{98.3} & 0.00 & +$\inf$ \\
    \;~LQ~(Lower Bound) & \cellcolor{SpringGreen!\SpringGreenP}{20.6} & \cellcolor{SpringGreen!\SpringGreenP}{38.9} & 249.71 & 19.75 \\
    \hline
    \;~SwinIR~\cite{sr_swinir} & \cellcolor{SpringGreen!\SpringGreenP}{22.6} & \cellcolor{SpringGreen!\SpringGreenP}{40.4} & 79.82 & \textbf{22.29} \\
    \;~SR4IR~\cite{kim2024beyond} & \cellcolor{SpringGreen!\SpringGreenP}{24.3} & \cellcolor{SpringGreen!\SpringGreenP}{42.1} & 79.17 & 22.01 \\
    \;~\textbf{NOLA-IR (Ours)} & \cellcolor{SpringGreen!\SpringGreenP}{\textbf{25.7}} & \cellcolor{SpringGreen!\SpringGreenP}{\textbf{43.3}} & \textbf{70.49} & 21.88 \\
    \bottomrule
    \end{tabular}
}
\vspace{-3mm}
\end{table}

\section{Conclusion}

In this paper, we identify stochasticity in diffusion-based restoration as a key obstacle for task-driven image restoration (TDIR) and propose NOLA, which repurposes a pretrained diffusion prior as a deterministic restoration module via a noise-free one-step forward pass with LoRA adaptation.
We show that the benefit of noise-free forwarding critically depends on the adaptation mechanism---LoRA yields consistent task-performance gains whereas ControlNet-style conditioning does not---and analyze this contrast through input-domain shift, task-feature recovery, and robustness to input perturbations.
To improve perceptual quality without compromising task performance, we further introduce task-preserving GAN training that better balances visual realism with task-driven objectives.
Extensive experiments on classification, segmentation, and detection show that NOLA-IR achieves state-of-the-art task performance while maintaining strong perceptual quality, and we further validate its generalization on real-world degraded images and OCR.
Overall, our findings provide practical guidance for leveraging pretrained diffusion priors in task-driven optimization and establish a strong baseline for future research on robust and generalizable TDIR.




\noindent \textbf{Acknowledgments.}
This work was supported in part by the IITPgrants [No. RS-2021-II211343, Artificial Intelligence Graduate School Program (Seoul National University), No. RS-2025-02303870, No.2022-0-00156] funded by the Korea government~(MSIT).

%
%
\bibliographystyle{splncs04}
\bibliography{main}

\clearpage

\begin{center}
    \vspace*{0.5cm}   
    {\Large\bfseries \emph{Supplementary Material for}\\[0.0em]
    Noise-Free One-Step LoRA for Task-Driven\\
    Image Restoration with Diffusion Priors\par}
    \vspace{1.0em}
    {\normalsize Jaeha Kim$^{1}$ \quad Kyoung Mu Lee$^{1,2}$\par}
    \vspace{0.5em}
    {\small
    $^{1}$Dept.\ of ECE \& ASRI, Seoul National University, Seoul, Korea\\
    $^{2}$IPAI, Seoul National University, Seoul, Korea\par}
    \vspace{0.3em}
    {\small \email{jhkim97s2@gmail.com, kyoungmu@snu.ac.kr}\par}
    \vspace{1.5em}
\end{center}

\appendix
\setcounter{section}{0}
\setcounter{figure}{0}
\setcounter{table}{0}
\setcounter{equation}{0}
\setcounter{algorithm}{0}

\renewcommand{\thetable}{S\arabic{table}}
\renewcommand{\thesection}{S\arabic{section}}
\renewcommand{\thefigure}{S\arabic{figure}}
\renewcommand{\theequation}{S\arabic{equation}}
\renewcommand{\thealgorithm}{S\arabic{algorithm}}

\renewcommand{\theHsection}{S\arabic{section}}
\renewcommand{\theHfigure}{S\arabic{figure}}
\renewcommand{\theHtable}{S\arabic{table}}
\renewcommand{\theHequation}{S\arabic{equation}}
\renewcommand{\theHalgorithm}{S\arabic{algorithm}}


In this supplementary material, we provide additional details, results, and ablation studies that are omitted from the main paper due to space constraints.
This document is organized as follows:

\begin{itemize}
\item \cref{sec:architecture_details}. Architecture Details
\item \cref{sec:additional_details}. Additional Implementation Details
\item \cref{sec:computational_cost}. Computational Cost
\item \cref{sec:additional_ablations}. Additional Ablations on Noise-Free Forward Pass
\item \cref{sec:quantitative_coco}. Quantitative Results: COCO-trained NOLA-IR
\item \cref{sec:performance_across}. Performance Across Different Tasks and Datasets
\item \cref{sec:visualization_main}. Additional Visualizations for the Main Benchmarks
\item \cref{sec:visualization_real_detection}. Additional Visualization for Real-World Detection
\item \cref{sec:visualization_ocr}. Additional Visualization for OCR
\end{itemize}

\section{Architecture Details}
\label{sec:architecture_details}

\cref{fig:nola} in the main manuscript illustrates the architecture used for NOLA-IR.
Given an input low-quality (LQ) image $x^{\mathrm{LQ}}$, we first obtain a pixel-error-optimized pre-restoration using a classic image restoration (IR) model $R_{\mathrm{pix}}$.
In our implementation, we use SwinIR~\cite{sr_swinir} as the pre-restoration module $R_{\mathrm{pix}}$ and train it with the pixel loss $\mathcal{L}_{\mathrm{pix}}=\|R_{\mathrm{pix}}(x^{\mathrm{LQ}})-x^{\mathrm{HQ}}\|_{1}$.

We then feed the pre-restored image into a pretrained Stable Diffusion (SD) model, denoted as $D_{\mathrm{SD}}$, which consists of a VAE encoder, a denoiser (denoising U-Net), and a VAE decoder.
All SD components are kept frozen, and we attach a trainable LoRA module (parameterized by $\theta$) only to the denoiser.

Finally, we apply wavelet-based color correction~\cite{wang2024exploiting} to preserve low-frequency components and prevent SD from altering the global appearance.
Overall, the noise-free one-step restoration in \cref{eq:one-step} can be written as
\begin{equation}
\hat{x}=\mathcal{G}_{\theta}(x^{\mathrm{LQ}})
=
\mathbf{W}_{\mathrm{high}}\!\Big(D_{\mathrm{SD}}\big(R_{\mathrm{pix}}(x^{\mathrm{LQ}})\big)\Big)
+
\mathbf{W}_{\mathrm{low}}\!\Big(R_{\mathrm{pix}}(x^{\mathrm{LQ}})\Big),
\label{eq:detailed-one-step}
\end{equation}
where $\mathbf{W}_{\mathrm{high}}(\cdot)$ and $\mathbf{W}_{\mathrm{low}}(\cdot)$ extract high- and low-frequency components via wavelet decomposition.

\section{Additional Implementation Details}
\label{sec:additional_details}

\subsubsection{Training NOLA-IR.}
\label{sssec:additional_details_training_nola-ir}
For training NOLA-IR, we use AdamW~\cite{misc_adamw} with a learning rate of $1\times10^{-4}$.
For the task networks (ResNet for classification, DeepLabV3 for segmentation, and Faster R-CNN for detection), we use SGD with a learning rate of $5\times10^{-3}$.
All learning rates are decayed using cosine annealing~\cite{misc_cosine_annealing} with $\eta_{\min}=1\times10^{-7}$.
We use a $512\times512$ input resolution and a batch size of 32/16/16 for classification/segmentation/detection, respectively.
We do not use text prompts for SD and instead use an empty prompt (\ie, \texttt{""}).
In \cref{eq:loss_G}, we set $\alpha=0.05$ and $\beta=0.1$, and we set $\gamma$ in \cref{eq:loss_F} to $1.0$ for classification, $0.5$ for segmentation, and $0.1$ for detection, following EDTR~\cite{kim2025exploiting}.
For $\mathcal{L}_{\mathrm{HLF}}$ and $\mathcal{L}_{\mathrm{FM}}$, we extract $\mathcal{H}^{\mathrm{feat}}_{\phi}$ from the feature map immediately after the backbone feature extractor (\eg, the output of MobileNetV3-Large in DeepLabV3), following SR4IR~\cite{kim2024beyond}.

\subsubsection{Batch-Wise Mixing.}
\label{sssec:additional_details_batch_mixing}
When updating the task network $\mathcal{H}_\phi$, we follow EDTR~\cite{kim2025exploiting} and construct each batch by concatenating restored and HQ images in equal proportion: for a batch size $B$,
\begin{equation}
x^{\mathrm{mix}} = \big[\hat{x}_{1:B/2};\, x^{\mathrm{HQ}}_{B/2+1:B}\big],
\label{eq:mix-supp}
\end{equation}
where $[\cdot\,;\cdot]$ denotes concatenation along the batch dimension.
This exposes $\mathcal{H}_\phi$ to both restored and clean images during training, stabilizing task-network training.

\subsubsection{Task-Preserving GAN Training.}
For task-preserving GAN training, we initialize from the task-driven LoRA checkpoint ($\mathcal{G}_{\theta^\star}$) and continue training with the adversarial and regularization losses using a halved learning rate.
To stabilize GAN training, we set $\alpha=0$ for the first 1k iterations to warm up the discriminator and provide meaningful gradients to the generator ($\mathcal{G}_\theta$).
For the discriminator $\mathcal{D}_\psi$, we use a frozen OpenCLIP visual encoder~\cite{cherti2023reproducible} with a lightweight trainable discriminator head, following~\cite{lin2025harnessing}.

\subsubsection{Compared Methods.}
We use the same task-network architectures for all compared methods in \cref{sssec:compared-methods}.
For the TDIR baselines, we use SwinIR~\cite{sr_swinir} as the IR backbone for TDSR~\cite{sr_tdsr}, RSRSSN~\cite{zhao2018residual}, and SR4IR~\cite{kim2024beyond}.
For DiffBIR~\cite{lin2023diffbir}, we disable classifier-free guidance (CFG)~\cite{ho2022classifier} since we use no prompt conditioning.

\subsubsection{Sensitivity to Input Perturbations.}
For the sensitivity analysis in \cref{ssec:abl-nola}, we perturb the input LQ image with additive Gaussian noise and measure the relative change of the denoiser output latent, $\|\mathcal{G}^{\mathrm{feat}}(x) - \mathcal{G}^{\mathrm{feat}}(x^{\mathrm{pert}})\|_2^2 / \|\mathcal{G}^{\mathrm{feat}}(x)\|_2^2$, where $\mathcal{G}^{\mathrm{feat}}$ denotes the output latent of the denoising U-Net.
We report the average over the segmentation validation set.


\subsubsection{OCR.}
\label{sssec:additional_details_training_ocr}
For the OCR experiments (\cref{ssec:main-paper-ocr}), we adopt the standard MJSynth~\cite{jaderberg2014synthetic} train/validation split (7.2M/0.8M), training on the full training set and reporting on a randomly sampled 10\% subset of the validation set for efficiency.
The CRNN~\cite{shi2016end}-based OCR network~\cite{baek2019wrong} uses a VGG backbone~\cite{net_vgg}.
We train for 150k iterations with Adam~\cite{misc_adam} (learning rate $5\times10^{-5}$, batch size 192).
\section{Computational Cost}
\label{sec:computational_cost}

\cref{table:computational-cost} compares the computational cost of SwinIR~\cite{sr_swinir}, DiffBIR~\cite{lin2023diffbir}, EDTR~\cite{kim2025exploiting}, and our NOLA-IR.
We report only the IR model cost, excluding the downstream task network, to focus on IR efficiency.
While NOLA-IR is heavier than the non-diffusion baseline SwinIR, it is substantially more efficient than prior SD-based methods in terms of trainable parameters, MACs, and inference time, thanks to parameter-efficient LoRA~\cite{hu2022lora} and one-step forwarding.

\vspace{-3mm}
\begin{table}[h!]
\caption{
Computational cost comparison of restoration models.
In \#Parameters, the \textcolor{gray}{gray} term denotes frozen parameters (mainly from the SD backbone), and the non-gray term denotes trainable parameters.
MACs and inference time are measured on a single NVIDIA A6000 GPU with a $512\times512$ input.
}
\small
\centering
\setlength\tabcolsep{1.0pt}
\def\arraystretch{1.1}
\label{table:computational-cost}
\resizebox{1.0\linewidth}{!}{
    \begin{tabular}{L{3.5cm}|C{3.3cm}|C{3.3cm}|C{3.3cm}}
    \toprule
    \;~Methods & \#Parameters (M) & MACs (G) & Inference Time (s) \\
    \hline
    \;~SwinIR~\cite{sr_swinir} & 15.8 & 86.9 & 0.055 \\
    \;~DiffBIR~\cite{lin2023diffbir} & \textcolor{gray}{1305.7} + 363.2 & 24338.5  & 4.824 \\
    \;~EDTR~\cite{kim2025exploiting} & \textcolor{gray}{1256.2} + 412.6 & 3689.1 & 0.525 \\
    \;~\textbf{NOLA-IR (Ours)} & \textcolor{gray}{1305.7} + 64.9 & 2333.0 & 0.219 \\
    \bottomrule
    \end{tabular}
}
\vspace{-3mm}
\end{table}
\section{Additional Ablations on Noise-Free Forward Pass}
\label{sec:additional_ablations}

\cref{table:nola-additional-ablation} provides additional evidence that the effect of the noise-free one-step forward pass depends on the adapter choice, consistent with the findings in \cref{ssec:abl-nola} of the main paper.

\subsubsection{Parameter-matched LoRA.}
We test a larger LoRA adapter (LoRA$_\mathrm{large}$) to rule out the possibility that the gains come simply from LoRA having fewer trainable parameters than ControlNet.
Specifically, we increase the LoRA rank from 64 to 360 so that LoRA$_\mathrm{large}$ has a comparable number of trainable parameters (364.9M) to ControlNet (363.2M).
Even with the enlarged LoRA, the noise-free forward pass consistently improves performance (about $+2\%$) across all tasks.
This supports our claim that the gains are driven by the adaptation mechanism (LoRA vs.\ ControlNet), rather than parameter count.

\subsubsection{Original ControlNet.}

While ``ControlNet'' in the main paper refers to IRControlNet~\cite{lin2023diffbir} (the ControlNet-style adapter in our SD-based backbone), here we verify that our observation is not specific to it: we replace IRControlNet with the original ControlNet~\cite{zhang2023adding} (denoted as ControlNet$^\dagger$) while keeping the SD backbone and training protocol unchanged.
As shown in \cref{table:nola-additional-ablation}, applying the noise-free forward pass still does not improve task performance, confirming that noise-free forwarding is not beneficial with ControlNet-style conditioning in our task-driven setting.

\begin{table}[t!]
\caption{
    Additional ablations on the noise-free one-step forward pass.
    LoRA$_\mathrm{large}$ and ControlNet$^\dagger$ denote the parameter-matched LoRA and the original ControlNet, respectively.
}
\vspace{-2mm}
\small
\centering
\setlength\tabcolsep{1.0pt}
\def\arraystretch{1.1}
\label{table:nola-additional-ablation}
\resizebox{0.95\linewidth}{!}{
    \begin{tabular}{L{5.5cm}|C{2.5cm}|C{2.5cm}|C{2.5cm}}
    \toprule
    \;~\multirow{2}{*}{Methods} & Classification & Segmentation & Detection \\
    & (Acc) & (mIoU) & (mAP) \\
    \hline
    \;~Baseline (LoRA$_\mathrm{large}$) & 67.2 & 60.4 & 29.8 \\
    \;~+Noise-Free Forward Pass & 69.8 \textcolor{green4mark}{(+2.6)} & 63.0 \textcolor{green4mark}{(+2.6)} & 31.9 \textcolor{green4mark}{(+2.1)} \\
    \hline
    \;~Baseline (ControlNet$^\dagger$) & 66.4 & 55.6 & 28.1 \\
    \;~+Noise-Free Forward Pass & 66.3 \textcolor{red4mark}{(-0.1)} & 54.8 \textcolor{red4mark}{(-0.8)} & 28.2 \textcolor{orange4mark}{(+0.1)} \\
    \bottomrule
    \end{tabular}
}
\vspace{-3mm}
\end{table}
\section{Quantitative Results: COCO-trained NOLA-IR}
\label{sec:quantitative_coco}

\cref{table:det-coco-table} reports object detection results on MS COCO 2017~\cite{data_coco}, following the setting in \cref{ssec:main-paper-real-world-det}: NOLA-IR is retrained on COCO and evaluated on the COCO 2017 validation set with synthetic Real-ESRGAN degradations.
Compared to SwinIR retrained under the same protocol, NOLA-IR substantially improves detection performance (mAP $16.4\rightarrow24.5$) while also achieving much better perceptual quality (\eg, FID $73.09\rightarrow11.91$), demonstrating that our approach generalizes to the large-scale COCO setting.

\begin{table}[h!]
\caption{
    Object detection performance and restoration quality on MS COCO 2017 validation with synthetic Real-ESRGAN degradations. \textbf{Bold} indicates the best result excluding the HQ upper-bound reference.
}
\small
\centering
\setlength\tabcolsep{1.0pt}
\def\arraystretch{1.1}
\label{table:det-coco-table}
\resizebox{0.9\linewidth}{!}{
    \begin{tabular}{L{4.0cm}|C{2.5cm}|C{1.8cm}C{1.8cm}C{1.8cm}}
    \toprule
    \,~\multirow{2}{*}{Settings} & \multicolumn{1}{c|}{Task} & \multicolumn{3}{c}{Visual quality} \\
    & mAP$_\uparrow$ & FID$_\downarrow$ & Q-Align$_\uparrow$ & PSNR$_\uparrow$ \\
    \hline
    \;~HQ~(Upper Bound) & \cellcolor{SpringGreen!\SpringGreenP}{44.1} & 0.00 & 4.11 & +$\inf$ \\
    \;~LQ~(Lower Bound) & \cellcolor{SpringGreen!\SpringGreenP}{11.7} & 124.12 & 1.03 & 20.28 \\
    \;~SwinIR~\cite{sr_swinir} & \cellcolor{SpringGreen!\SpringGreenP}{16.4} & 73.09 & 1.90 & \textbf{21.22} \\
    \;~\textbf{NOLA-IR (Ours)} & \cellcolor{SpringGreen!\SpringGreenP}{\textbf{24.5}} & \textbf{11.91} & \textbf{4.05} & 19.58 \\
    \bottomrule
    \end{tabular}
}
\vspace{-3mm}
\end{table}

\section{Performance Across Different Tasks and Datasets}
\label{sec:performance_across}
To further examine generalization, we evaluate NOLA-IR across (i) different downstream tasks and (ii) different datasets.

\subsubsection{Performance Across Different Tasks.}
\cref{table:performance-across-tasks} reports cross-task evaluation across classification, segmentation, and detection.
Here, $\mathcal{G}_\theta^{(task)}$ denotes the NOLA-IR model trained with task $task$.
We swap only $\mathcal{G}_\theta$ while keeping the corresponding task network fixed, to test cross-task transfer of the IR model.

Each model performs best on the task it was trained for (\ie, diagonal entries).
Notably, $\mathcal{G}_\theta^{(\text{Segmentation})}$ and $\mathcal{G}_\theta^{(\text{Detection})}$ transfer reasonably well to each other, whereas $\mathcal{G}_\theta^{(\text{Classification})}$ generalizes poorly to segmentation and detection, likely due to the fine-grained nature of CUB-200-2011~\cite{WahCUB_200_2011}, which biases restoration toward subtle category-specific (\ie, bird-specific) textures.
\begin{table}[h!]
\caption{
    Cross-task evaluation of NOLA-IR. \textbf{Bold} indicates the best result in each evaluation task (column-wise).
}
\small
\centering
\setlength\tabcolsep{1.0pt}
\def\arraystretch{1.1}
\label{table:performance-across-tasks}
\resizebox{0.90\linewidth}{!}{
    \begin{tabular}{L{3.5cm}|C{2.4cm}|C{2.4cm}|C{2.4cm}}
    \toprule
    \multirow{2}{*}{\diagbox[width=3.570cm]{\;~$\mathcal{G}_\theta^{(task)}$}{Eval Task\;\;}} & Classification & Segmentation & Detection \\
    & (Acc$_\uparrow$) & (mIoU$_\uparrow$) & (mAP$_\uparrow$) \\
    \hline
    \;~$\mathcal{G}_\theta^\text{(Classification)}$ & \textbf{69.3} & 47.9 & 21.0 \\
    \;~$\mathcal{G}_\theta^\text{(Segmentation)}$ & 59.8 & \textbf{62.9} & 29.9 \\
    \;~$\mathcal{G}_\theta^\text{(Detection)}$ & 60.8 & 61.7 & \textbf{32.0} \\
    \bottomrule
    \end{tabular}
}
\vspace{-3mm}
\end{table}

\vspace{-3mm}
\subsubsection{Performance Across Different Datasets.}
\cref{table:performance-across-datasets} reports cross-dataset generalization results.
We train IR models on degraded PASCAL VOC 2012 (our main benchmark setting) and evaluate them on degraded MS COCO 2017, where degradations are synthesized using the same Real-ESRGAN pipeline.
For evaluation on COCO, we use a detector trained on COCO HQ images and feed it restored images produced by each IR method.

As shown in \cref{table:performance-across-datasets}, NOLA-IR achieves the best mAP on the unseen COCO dataset among compared restoration methods, demonstrating stronger cross-dataset generalization.

\begin{table}[b!]
\vspace{-3mm}
\caption{
    Cross-dataset evaluation of IR models for object detection (mAP). \textbf{Bold} indicates the best result excluding the HQ upper-bound reference.
}
\small
\centering
\setlength\tabcolsep{1.0pt}
\def\arraystretch{1.1}
\label{table:performance-across-datasets}
\resizebox{0.95\linewidth}{!}{
    \begin{tabular}{L{4.0cm}|C{3.8cm}|C{3.8cm}}
    \toprule
    \multirow{2}{*}{\diagbox[width=4.07cm]{\;~Settings}{Eval Dataset\;\;}}  &  \textbf{\textcolor{teal}{Seen}} & \multicolumn{1}{c}{\textbf{\textcolor{purple}{Unseen}}} \\
    & PASCAL VOC 2012~\cite{everingham2010pascal} & MS COCO 2017~\cite{data_coco} \\
    \hline
    \;~HQ (Upper Bound) & 37.1 & 44.1 \\
    \hline
    \;~SwinIR~\cite{sr_swinir} & 19.4 & 7.4 \\
    \;~DiffBIR~\cite{lin2023diffbir} & 22.8 & 10.2 \\
    \;~TDSR~\cite{sr_tdsr} & 13.1 & 8.2 \\
    \;~RSRSSN~\cite{zhao2018residual} & 19.9 & 7.6 \\
    \;~SR4IR~\cite{kim2024beyond} & 20.0 & 8.3 \\
    \;~EDTR~\cite{kim2025exploiting} & 30.2 & 16.2 \\
    \;~\textbf{NOLA-IR (Ours)} & \textbf{32.0} & \textbf{19.7} \\
    \bottomrule
    \end{tabular}
}
\end{table}

\section{Additional Visualizations for the Main Benchmarks}
\label{sec:visualization_main}
\begin{figure*}[b!]
    \vspace{-3mm}
    \centering
    \captionsetup[subfigure]{labelfont=scriptsize, textfont=scriptsize}
    \renewcommand{\wp}{0.240}
    \renewcommand{\splitT}{0.250}

    \begin{tikzpicture}[remember picture]
        \node[inner sep=0pt] (grid) {%
            \begin{minipage}{\textwidth}
                \centering
                \subfloat[LQ~(Lower Bound)]{\includegraphics[width=\wp\linewidth]{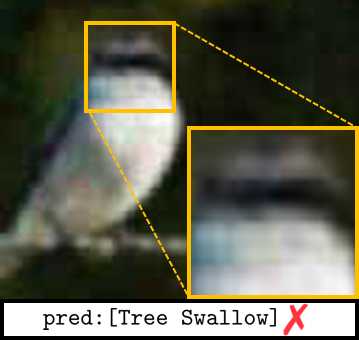}}
                \vspace{1mm}
                \hfill
                \subfloat[SwinIR~\cite{sr_swinir}]{\includegraphics[width=\wp\linewidth]{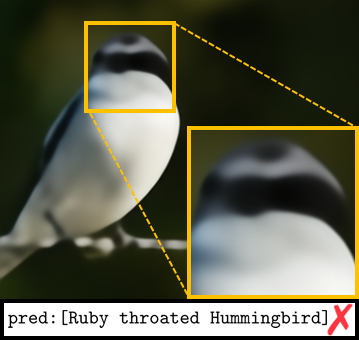}}
                \hfill
                \subfloat[DiffBIR~\cite{lin2023diffbir}]{\includegraphics[width=\wp\linewidth]{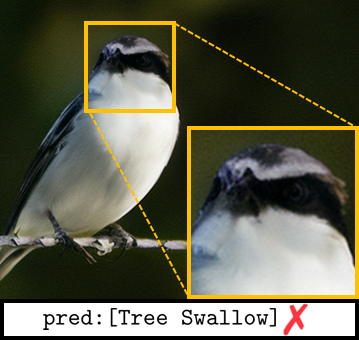}}
                \hfill
                \subfloat[SR4IR~\cite{kim2024beyond}]{\includegraphics[width=\wp\linewidth]{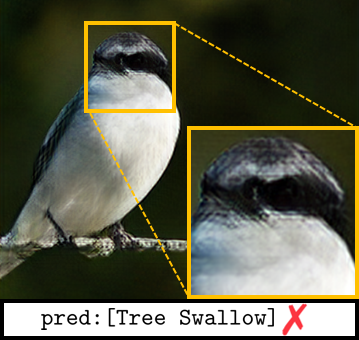}}

                \par\vspace{2mm}

                \subfloat[HQ~(Ground-truth)]{\includegraphics[width=\wp\linewidth]{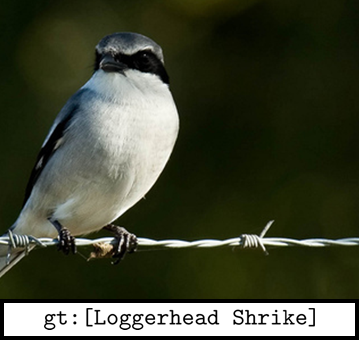}}
                \vspace{1mm}
                \hfill
                \subfloat[EDTR~\cite{kim2025exploiting}]{\includegraphics[width=\wp\linewidth]{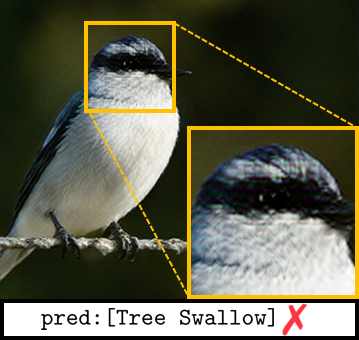}}
                \hfill
                \subfloat[\textbf{NOLA-IR}]{\includegraphics[width=\wp\linewidth]{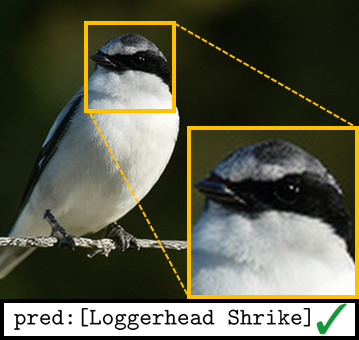}}
                \hfill
                \subfloat[HQ~(Upper Bound)]{\includegraphics[width=\wp\linewidth]{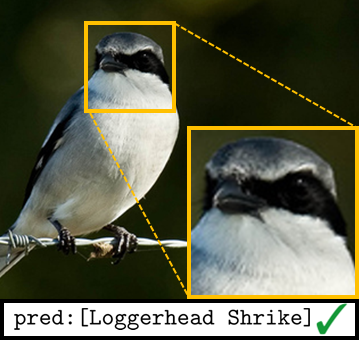}}
            \end{minipage}%
        };

        \begin{scope}[overlay]
            \draw[dashed, line width=0.8pt]
              ($(grid.north west)!\splitT!(grid.north east)$) -- ($(grid.south west)!\splitT!(grid.south east)$);
        \end{scope}
    \end{tikzpicture}

    \addtocounter{subfigure}{-8}
    \vspace{8mm}

    \begin{tikzpicture}[remember picture]
        \node[inner sep=0pt] (grid) {%
            \begin{minipage}{\textwidth}
                \centering
                \subfloat[LQ~(Lower Bound)]{\includegraphics[width=\wp\linewidth]{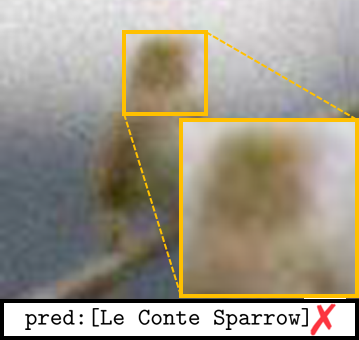}}
                \vspace{1mm}
                \hfill
                \subfloat[SwinIR~\cite{sr_swinir}]{\includegraphics[width=\wp\linewidth]{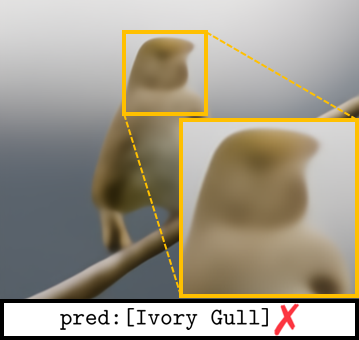}}
                \hfill
                \subfloat[DiffBIR~\cite{lin2023diffbir}]{\includegraphics[width=\wp\linewidth]{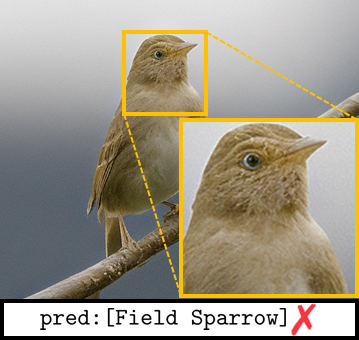}}
                \hfill
                \subfloat[SR4IR~\cite{kim2024beyond}]{\includegraphics[width=\wp\linewidth]{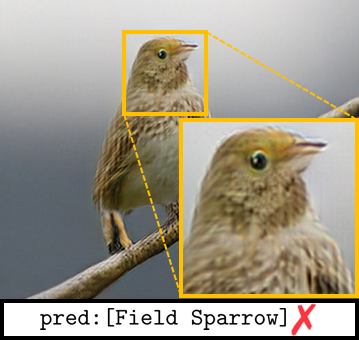}}

                \par\vspace{2mm}

                \subfloat[HQ~(Ground-truth)]{\includegraphics[width=\wp\linewidth]{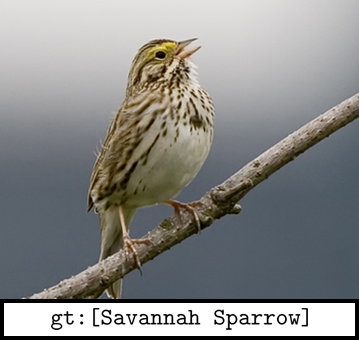}}
                \vspace{1mm}
                \hfill
                \subfloat[EDTR~\cite{kim2025exploiting}]{\includegraphics[width=\wp\linewidth]{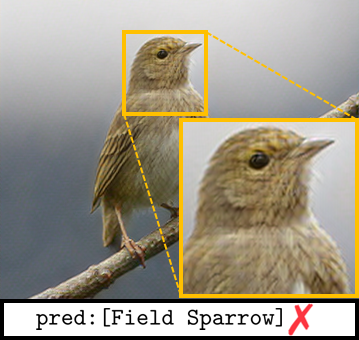}}
                \hfill
                \subfloat[\textbf{NOLA-IR}]{\includegraphics[width=\wp\linewidth]{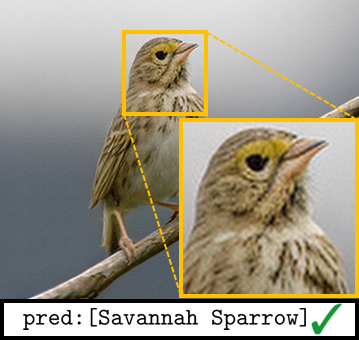}}
                \hfill
                \subfloat[HQ~(Upper Bound)]{\includegraphics[width=\wp\linewidth]{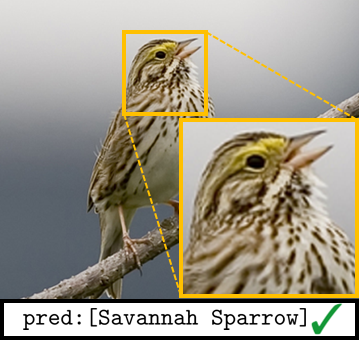}}
            \end{minipage}%
        };

        \begin{scope}[overlay]
            \draw[dashed, line width=0.8pt]
              ($(grid.north west)!\splitT!(grid.north east)$) -- ($(grid.south west)!\splitT!(grid.south east)$);
        \end{scope}
    \end{tikzpicture}

    \addtocounter{subfigure}{-8}

    \caption{Additional qualitative results on CUB-200-2011 with downstream classification outcomes. Notation follows \cref{fig:vis-cls} in the main paper.}
    \label{fig:additional-vis-cls}
\end{figure*}

\begin{figure*}[t!]
    \centering
    \captionsetup[subfigure]{labelfont=scriptsize, textfont=scriptsize}
    \renewcommand{\wp}{0.240}
    \renewcommand{\splitT}{0.250}

    \begin{tikzpicture}[remember picture]
        \node[inner sep=0pt] (grid) {%
            \begin{minipage}{\textwidth}
                \centering
                \subfloat[LQ (Lower Bound)]{\includegraphics[width=\wp\linewidth]{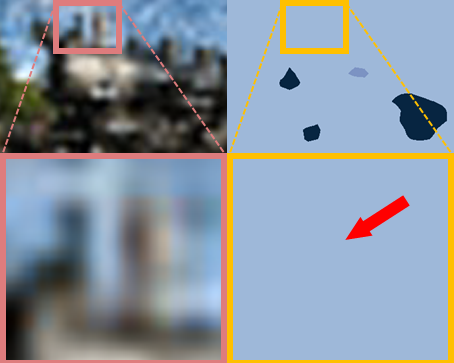}}
                \vspace{1mm}
                \hfill
                \subfloat[SwinIR~\cite{sr_swinir}]{\includegraphics[width=\wp\linewidth]{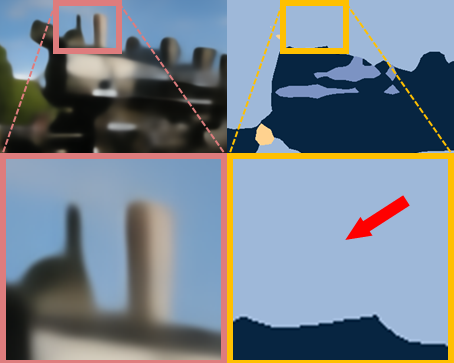}}
                \hfill
                \subfloat[DiffBIR~\cite{lin2023diffbir}]{\includegraphics[width=\wp\linewidth]{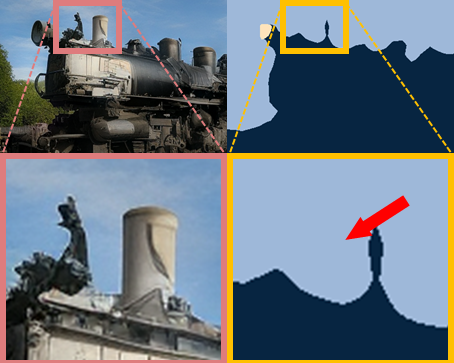}}
                \hfill
                \subfloat[SR4IR~\cite{kim2024beyond}]{\includegraphics[width=\wp\linewidth]{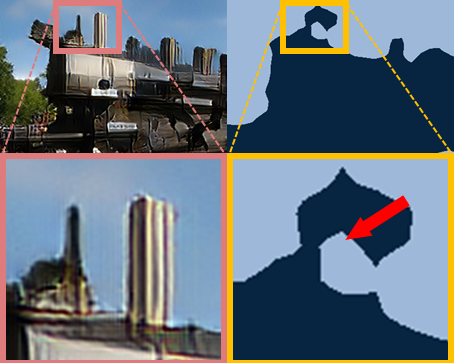}}


                \subfloat[HQ~(Ground-truth)]{\includegraphics[width=\wp\linewidth]{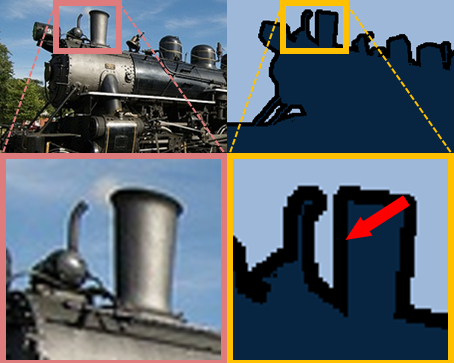}}
                \vspace{1mm}
                \hfill
                \subfloat[EDTR~\cite{kim2025exploiting}]{\includegraphics[width=\wp\linewidth]{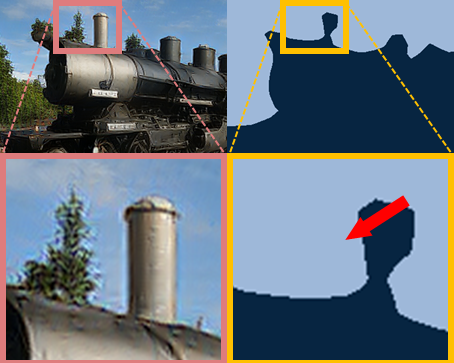}}
                \hfill
                \subfloat[\textbf{NOLA-IR}]{\includegraphics[width=\wp\linewidth]{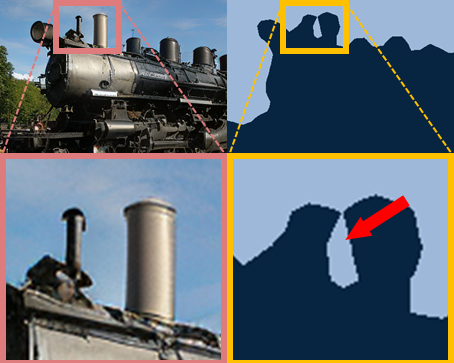}}
                \hfill
                \subfloat[HQ~(Upper Bound)]{\includegraphics[width=\wp\linewidth]{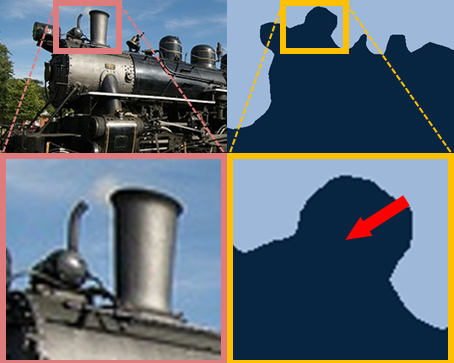}}
            \end{minipage}%
        };

        \begin{scope}[overlay]
            \draw[dashed, line width=0.8pt]
              ($(grid.north west)!\splitT!(grid.north east)$) -- ($(grid.south west)!\splitT!(grid.south east)$);
        \end{scope}
    \end{tikzpicture}

    \vspace{10mm}
    \addtocounter{subfigure}{-8}

    \begin{tikzpicture}[remember picture]
        \node[inner sep=0pt] (grid) {%
            \begin{minipage}{\textwidth}
                \centering
                \subfloat[LQ (Lower Bound)]{\includegraphics[width=\wp\linewidth]{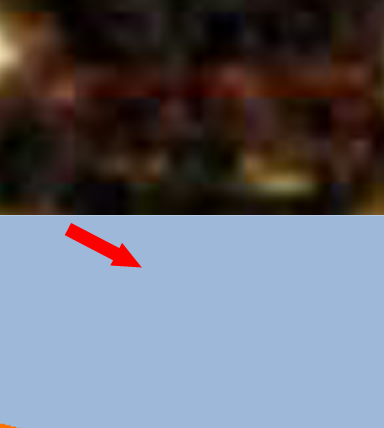}}
                \vspace{1mm}
                \hfill
                \subfloat[SwinIR~\cite{sr_swinir}]{\includegraphics[width=\wp\linewidth]{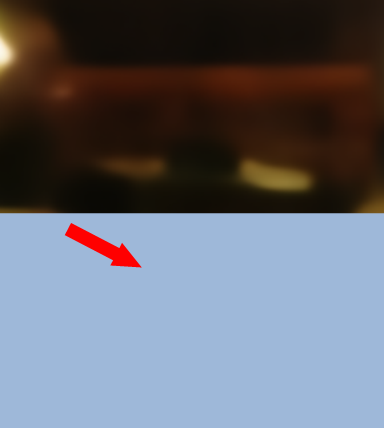}}
                \hfill
                \subfloat[DiffBIR~\cite{lin2023diffbir}]{\includegraphics[width=\wp\linewidth]{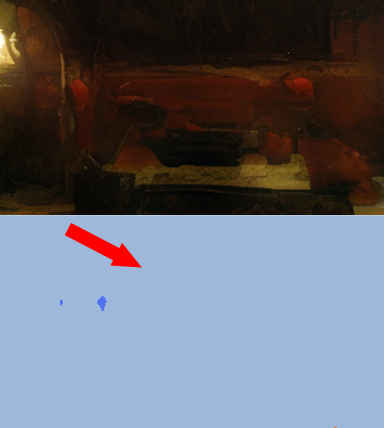}}
                \hfill
                \subfloat[SR4IR~\cite{kim2024beyond}]{\includegraphics[width=\wp\linewidth]{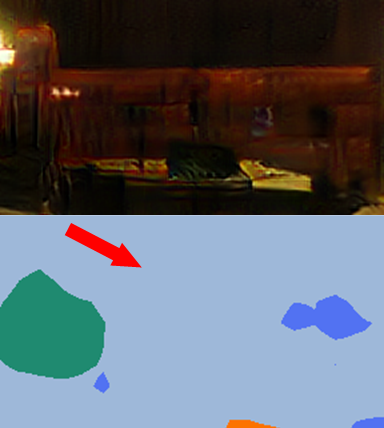}}


                \subfloat[HQ~(Ground-truth)]{\includegraphics[width=\wp\linewidth]{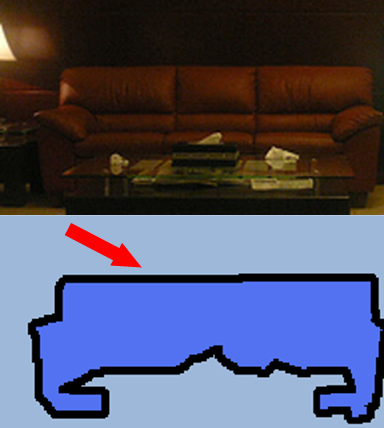}}
                \vspace{1mm}
                \hfill
                \subfloat[EDTR~\cite{kim2025exploiting}]{\includegraphics[width=\wp\linewidth]{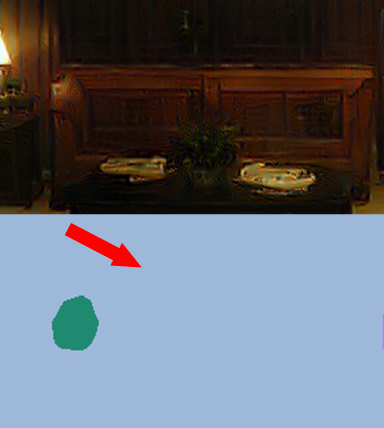}}
                \hfill
                \subfloat[\textbf{NOLA-IR}]{\includegraphics[width=\wp\linewidth]{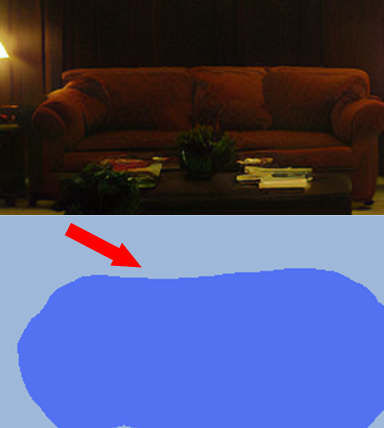}}
                \hfill
                \subfloat[HQ~(Upper Bound)]{\includegraphics[width=\wp\linewidth]{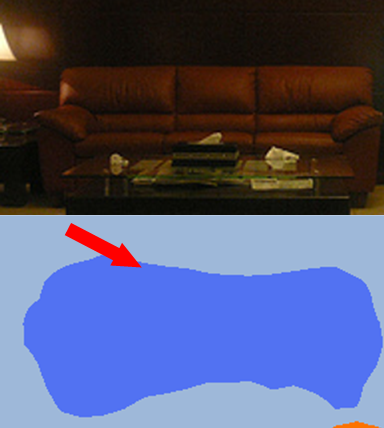}}
            \end{minipage}%
        };

        \begin{scope}[overlay]
            \draw[dashed, line width=0.8pt]
              ($(grid.north west)!\splitT!(grid.north east)$) -- ($(grid.south west)!\splitT!(grid.south east)$);
        \end{scope}
    \end{tikzpicture}

    \addtocounter{subfigure}{-8}

    \caption{Additional qualitative results on PASCAL VOC 2012 with downstream segmentation outcomes. Notation follows \cref{fig:vis-cls} in the main paper. In the ground-truth segmentation map (e), black denotes the ``Don't Care'' region.}
    \label{fig:additional-vis-seg}
    \vspace{-2mm}
\end{figure*}

\begin{figure*}[t!]
    \centering
    \captionsetup[subfigure]{labelfont=scriptsize, textfont=scriptsize}
    \renewcommand{\wp}{0.240}
    \renewcommand{\splitT}{0.250}

    \begin{tikzpicture}[remember picture]
        \node[inner sep=0pt] (grid) {%
            \begin{minipage}{\textwidth}
                \centering
                \subfloat[LQ (Lower Bound)]{\includegraphics[width=\wp\linewidth]{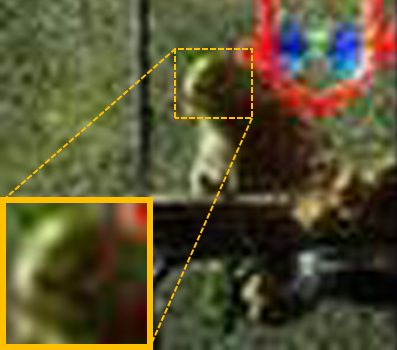}}
                \vspace{1mm}
                \hfill
                \subfloat[SwinIR~\cite{sr_swinir}]{\includegraphics[width=\wp\linewidth]{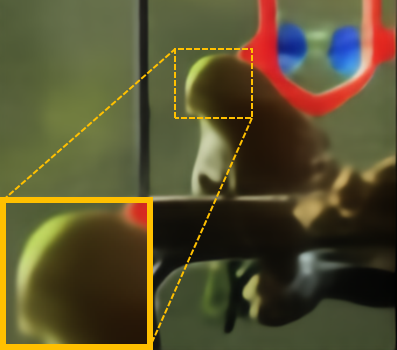}}
                \hfill
                \subfloat[DiffBIR~\cite{lin2023diffbir}]{\includegraphics[width=\wp\linewidth]{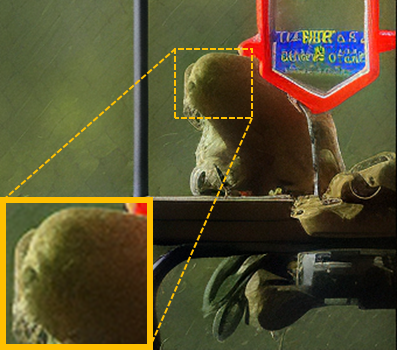}}
                \hfill
                \subfloat[SR4IR~\cite{kim2024beyond}]{\includegraphics[width=\wp\linewidth]{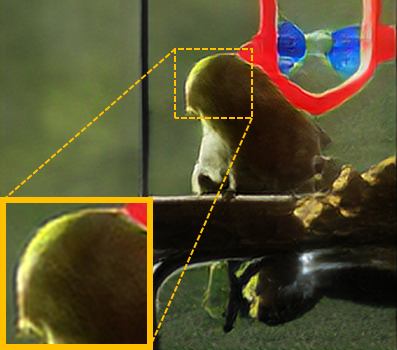}}

                \par\vspace{2mm}

                \subfloat[HQ~(Ground-truth)]{\includegraphics[width=\wp\linewidth]{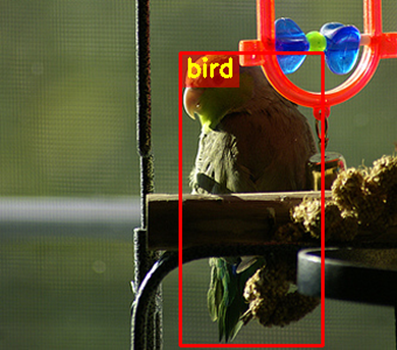}}
                \vspace{1mm}
                \hfill
                \subfloat[EDTR~\cite{kim2025exploiting}]{\includegraphics[width=\wp\linewidth]{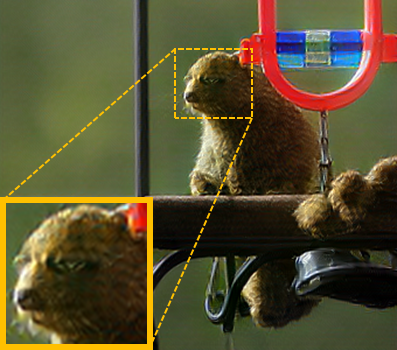}}
                \hfill
                \subfloat[\textbf{NOLA-IR}]{\includegraphics[width=\wp\linewidth]{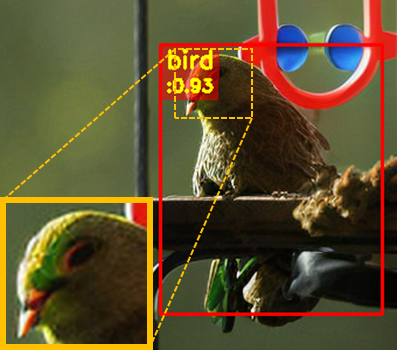}}
                \hfill
                \subfloat[HQ~(Upper Bound)]{\includegraphics[width=\wp\linewidth]{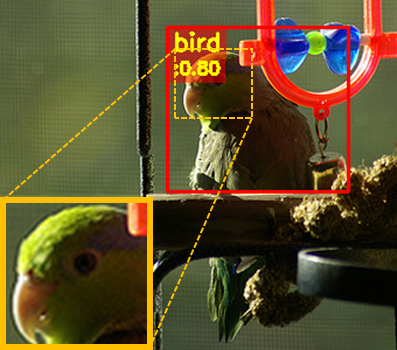}}
            \end{minipage}%
        };

        \begin{scope}[overlay]
            \draw[dashed, line width=0.8pt]
              ($(grid.north west)!\splitT!(grid.north east)$) -- ($(grid.south west)!\splitT!(grid.south east)$);
        \end{scope}
    \end{tikzpicture}

    \vspace{8mm}

    \addtocounter{subfigure}{-8}

    \begin{tikzpicture}[remember picture]
        \node[inner sep=0pt] (grid) {%
            \begin{minipage}{\textwidth}
                \centering
                \subfloat[LQ (Lower Bound)]{\includegraphics[width=\wp\linewidth]{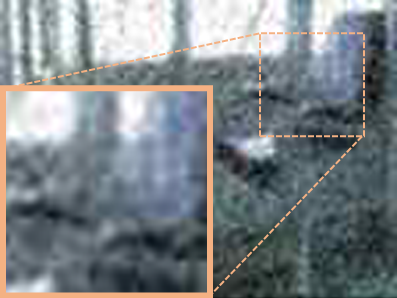}}
                \vspace{1mm}
                \hfill
                \subfloat[SwinIR~\cite{sr_swinir}]{\includegraphics[width=\wp\linewidth]{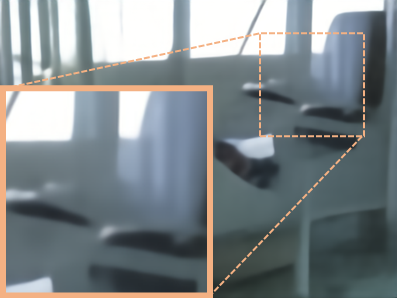}}
                \hfill
                \subfloat[DiffBIR~\cite{lin2023diffbir}]{\includegraphics[width=\wp\linewidth]{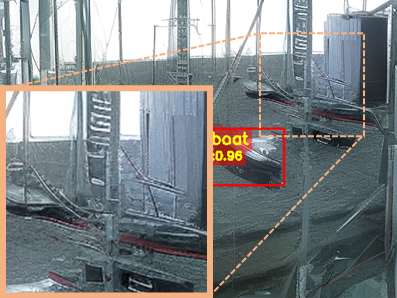}}
                \hfill
                \subfloat[SR4IR~\cite{kim2024beyond}]{\includegraphics[width=\wp\linewidth]{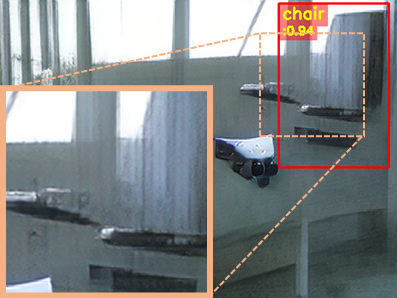}}

                \par\vspace{2mm}

                \subfloat[HQ~(Ground-truth)]{\includegraphics[width=\wp\linewidth]{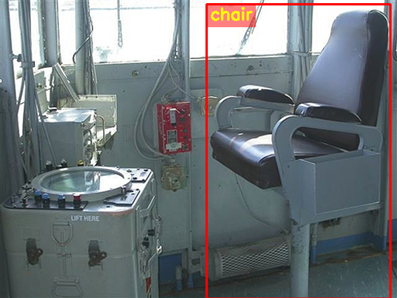}}
                \vspace{1mm}
                \hfill
                \subfloat[EDTR~\cite{kim2025exploiting}]{\includegraphics[width=\wp\linewidth]{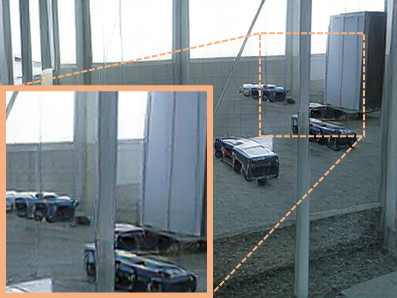}}
                \hfill
                \subfloat[\textbf{NOLA-IR}]{\includegraphics[width=\wp\linewidth]{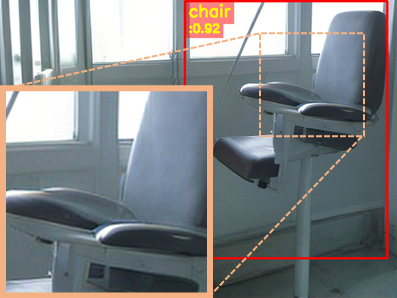}}
                \hfill
                \subfloat[HQ~(Upper Bound)]{\includegraphics[width=\wp\linewidth]{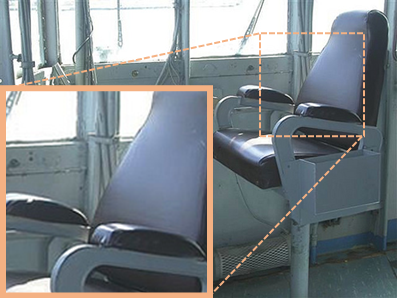}}
            \end{minipage}%
        };

        \begin{scope}[overlay]
            \draw[dashed, line width=0.8pt]
              ($(grid.north west)!\splitT!(grid.north east)$) -- ($(grid.south west)!\splitT!(grid.south east)$);
        \end{scope}
    \end{tikzpicture}

    \addtocounter{subfigure}{-8}

    \caption{Additional qualitative results on PASCAL VOC 2012 with downstream detection outcomes. Notation follows \cref{fig:vis-cls} in the main paper.}
    \label{fig:additional-vis-det}
\end{figure*}

\cref{fig:additional-vis-cls,fig:additional-vis-seg,fig:additional-vis-det} provide additional qualitative results for classification, segmentation, and detection on the main benchmarks.
Overall, NOLA-IR more consistently restores task-relevant details and yields more accurate predictions than the compared methods, consistent with the results in the main paper.

\clearpage

\clearpage
\section{Additional Visualization for Real-World Detection}
\label{sec:visualization_real_detection}

\begin{figure*}[t!]
    \centering
    \captionsetup[subfigure]{labelfont=scriptsize, textfont=scriptsize}
    \renewcommand{\wp}{0.325}
    \renewcommand{\splitT}{0.250}
    \resizebox{1.0\linewidth}{!}{
        \begin{minipage}{\textwidth}
            \centering
            \addtocounter{subfigure}{-3}
            \subfloat{\includegraphics[width=\wp\linewidth]{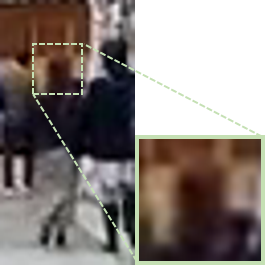}}
            \hfill
            \subfloat{\includegraphics[width=\wp\linewidth]{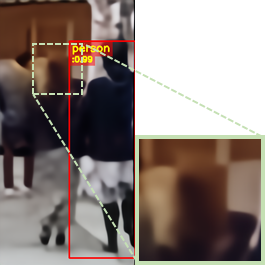}}
            \hfill
            \subfloat{\includegraphics[width=\wp\linewidth]{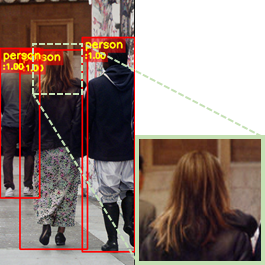}}
            \\
            \vspace{3mm}
            \addtocounter{subfigure}{-3}
            \subfloat{\includegraphics[width=\wp\linewidth]{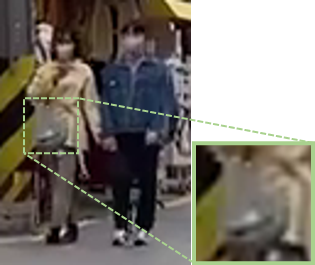}}
            \hfill
            \subfloat{\includegraphics[width=\wp\linewidth]{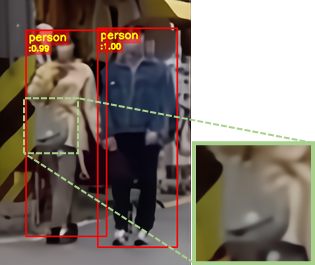}}
            \hfill
            \subfloat{\includegraphics[width=\wp\linewidth]{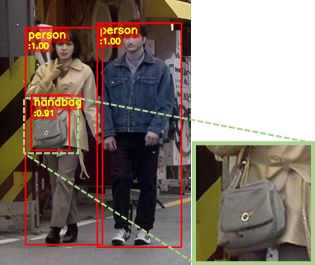}}
            \\
            \vspace{3mm}
            \addtocounter{subfigure}{-3}
            \subfloat{\includegraphics[width=\wp\linewidth]{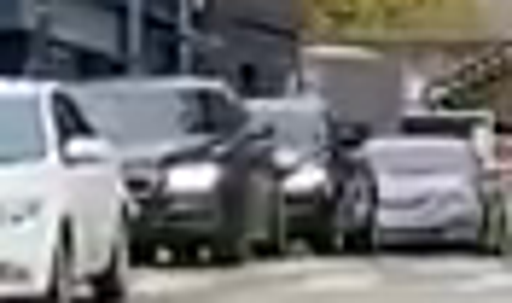}}
            \hfill
            \subfloat{\includegraphics[width=\wp\linewidth]{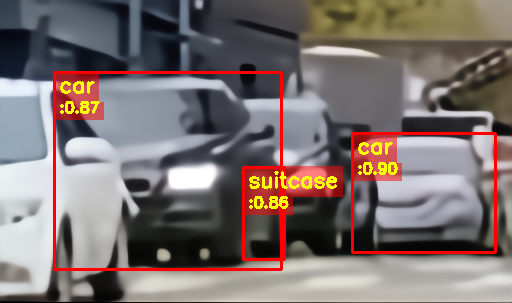}}
            \hfill
            \subfloat{\includegraphics[width=\wp\linewidth]{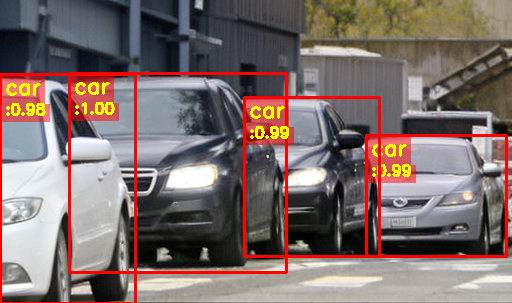}}
            \\
            \vspace{3mm}
            \addtocounter{subfigure}{-3}
            \subfloat{\includegraphics[width=\wp\linewidth]{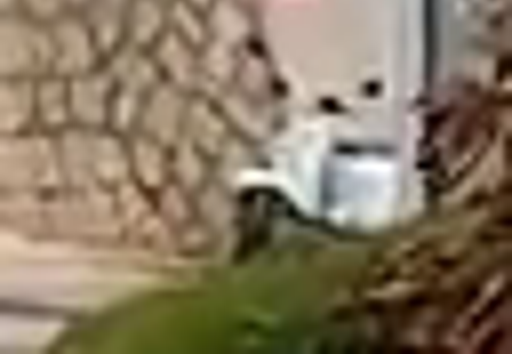}}
            \hfill
            \subfloat{\includegraphics[width=\wp\linewidth]{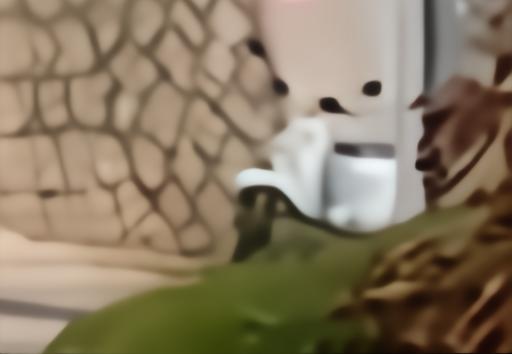}}
            \hfill
            \subfloat{\includegraphics[width=\wp\linewidth]{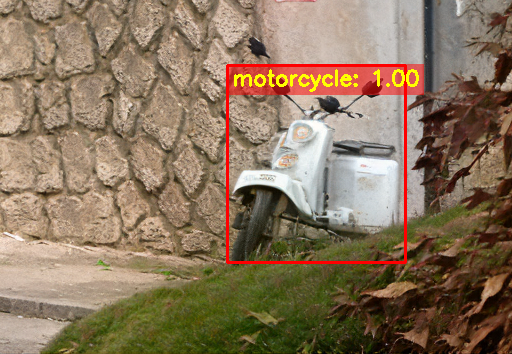}}
            \\
            \vspace{3mm}
            \addtocounter{subfigure}{-3}
            \subfloat[LQ]{\includegraphics[width=\wp\linewidth]{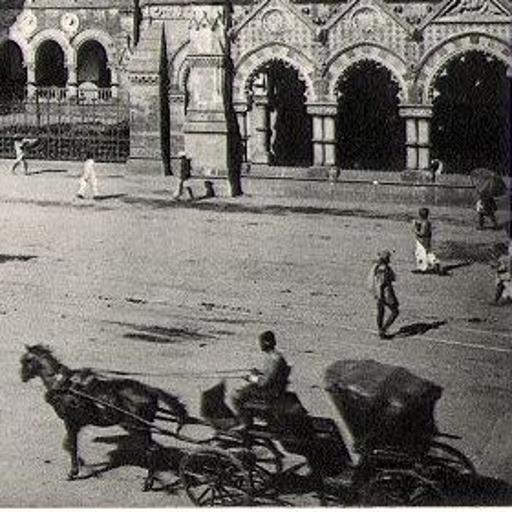}}
            \hfill
            \subfloat[SwinIR~\cite{sr_swinir}]{\includegraphics[width=\wp\linewidth]{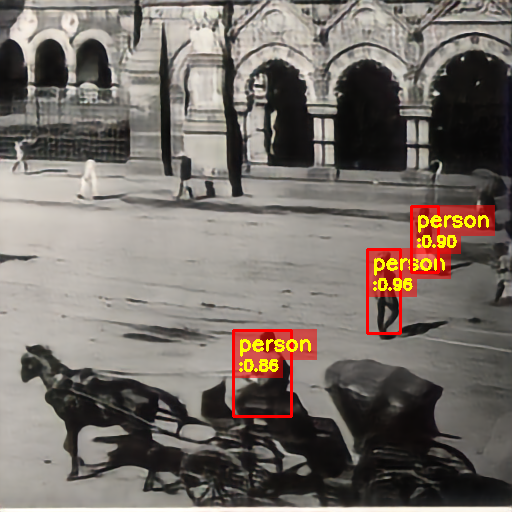}}
            \hfill
            \subfloat[\textbf{NOLA-IR}]{\includegraphics[width=\wp\linewidth]{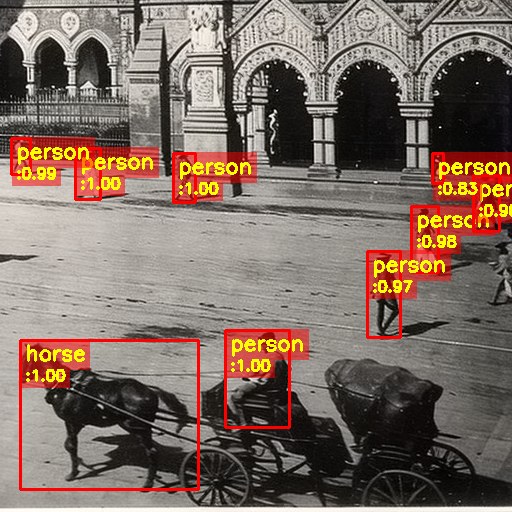}}
        \end{minipage}%
        \addtocounter{subfigure}{-3}
    }
    \caption{
    Restoration results and detection outcomes on real-world LQ inputs. Notation follows \cref{fig:vis-real-world-det} in the main paper. Rows 1--4 show small patches cropped from YouTube video frames, and the last row shows an old-photo image crawled from the Internet.
    }
    \label{fig:vis-real-world-det1}
\end{figure*}

\begin{figure*}[t!]
    \centering
    \captionsetup[subfigure]{labelfont=scriptsize, textfont=scriptsize}
    \renewcommand{\wp}{0.325}
    \renewcommand{\splitT}{0.250}
    \resizebox{1.0\linewidth}{!}{
        \begin{minipage}{\textwidth}
            \centering
            \addtocounter{subfigure}{-3}
            \subfloat{\includegraphics[width=\wp\linewidth]{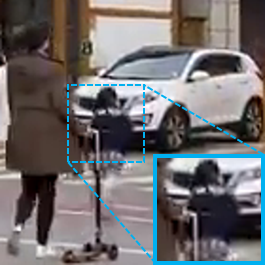}}
            \hfill
            \subfloat{\includegraphics[width=\wp\linewidth]{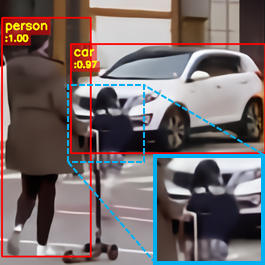}}
            \hfill
            \subfloat{\includegraphics[width=\wp\linewidth]{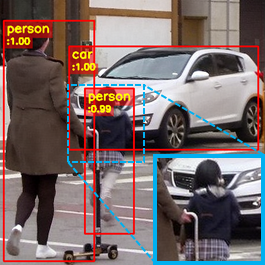}}
            \\
            \vspace{3mm}
            \addtocounter{subfigure}{-3}
            \subfloat{\includegraphics[width=\wp\linewidth]{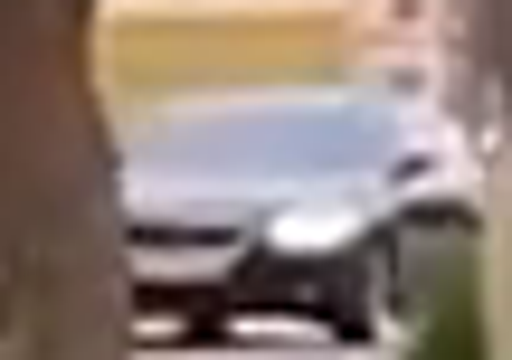}}
            \hfill
            \subfloat{\includegraphics[width=\wp\linewidth]{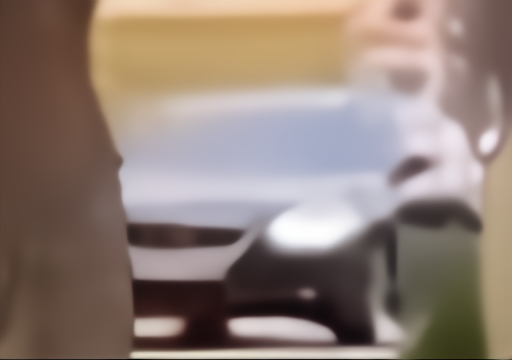}}
            \hfill
            \subfloat{\includegraphics[width=\wp\linewidth]{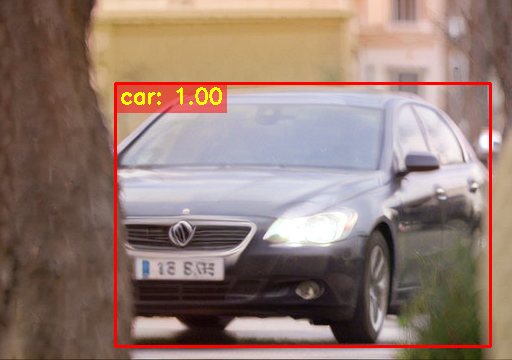}}
            \\
            \vspace{3mm}
            \addtocounter{subfigure}{-3}
            \subfloat{\includegraphics[width=\wp\linewidth]{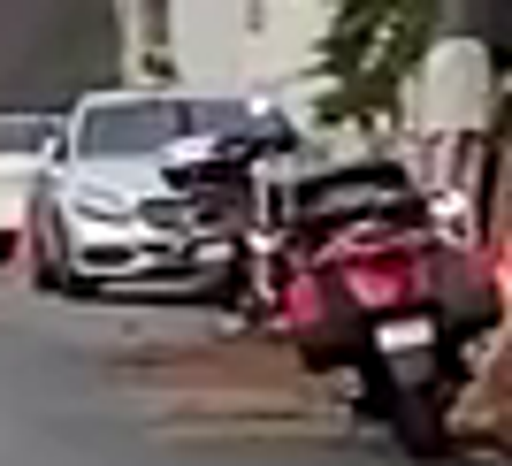}}
            \hfill
            \subfloat{\includegraphics[width=\wp\linewidth]{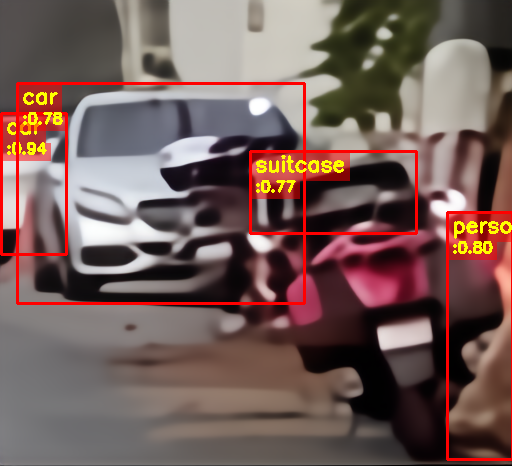}}
            \hfill
            \subfloat{\includegraphics[width=\wp\linewidth]{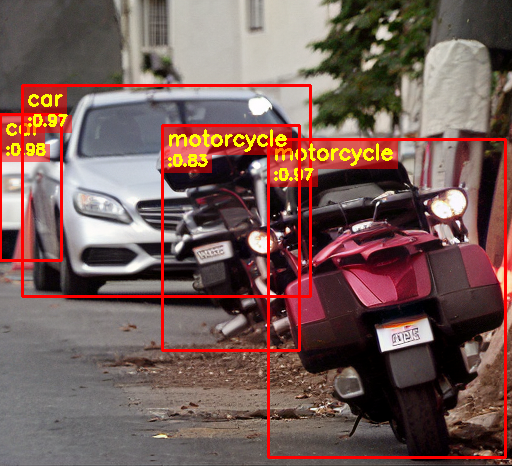}}
            \\
            \vspace{3mm}
            \addtocounter{subfigure}{-3}
            \subfloat{\includegraphics[width=\wp\linewidth]{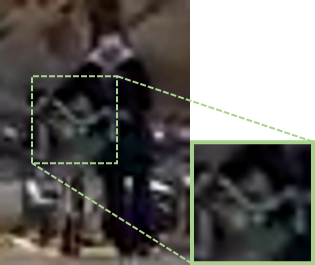}}
            \hfill
            \subfloat{\includegraphics[width=\wp\linewidth]{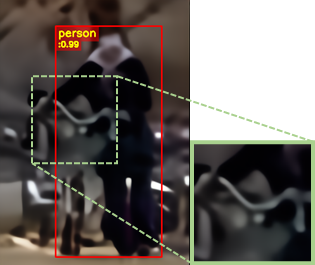}}
            \hfill
            \subfloat{\includegraphics[width=\wp\linewidth]{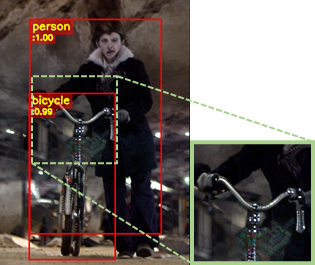}}
            \\
            \vspace{3mm}
            \addtocounter{subfigure}{-3}
            \subfloat[LQ]{\includegraphics[width=\wp\linewidth]{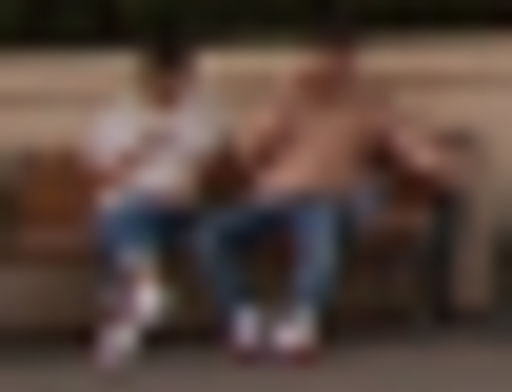}}
            \hfill
            \subfloat[SwinIR~\cite{sr_swinir}]{\includegraphics[width=\wp\linewidth]{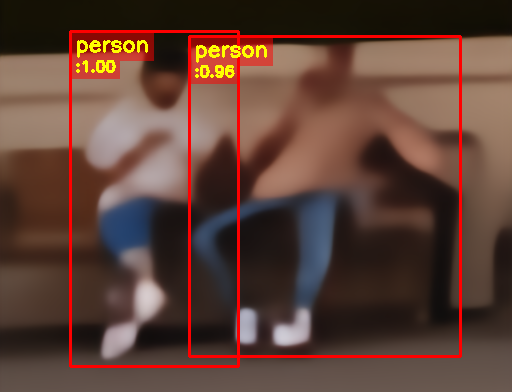}}
            \hfill
            \subfloat[\textbf{NOLA-IR}]{\includegraphics[width=\wp\linewidth]{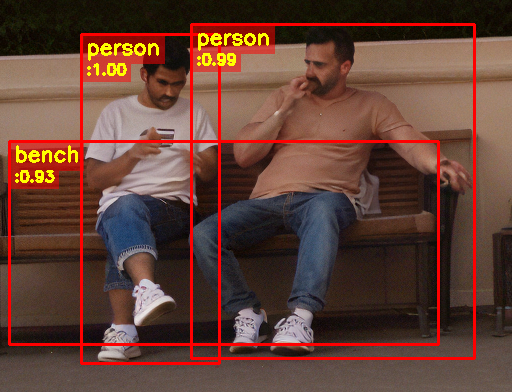}}
        \end{minipage}%
        \addtocounter{subfigure}{-3}
    }
    \caption{
    Restoration results and detection outcomes on real-world LQ inputs. Notation follows \cref{fig:vis-real-world-det} in the main paper. All examples are small patches cropped from YouTube video frames.
    }
    \label{fig:vis-real-world-det2}
\end{figure*}

\begin{figure*}[t!]
    \centering
    \captionsetup[subfigure]{labelfont=scriptsize, textfont=scriptsize}
    \renewcommand{\wp}{0.33}
    \renewcommand{\splitT}{0.250}
    \resizebox{1.0\linewidth}{!}{
        \begin{minipage}{\textwidth}
            \centering
            \addtocounter{subfigure}{-3}
            \subfloat{\includegraphics[width=0.32\linewidth]{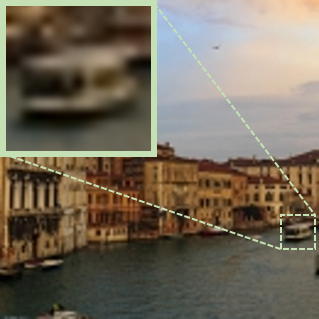}}
            \hfill
            \subfloat{\includegraphics[width=0.32\linewidth]{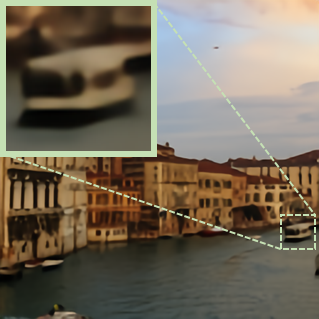}}
            \hfill
            \subfloat{\includegraphics[width=0.32\linewidth]{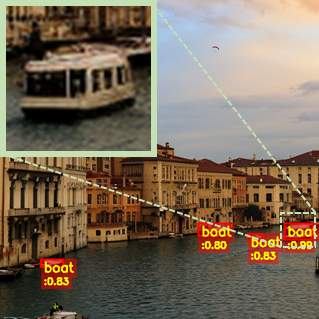}}
            \\
            \vspace{3mm}
            \addtocounter{subfigure}{-3}
            \subfloat{\includegraphics[width=0.32\linewidth]{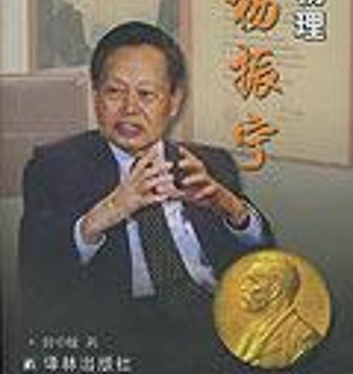}}
            \hfill
            \subfloat{\includegraphics[width=0.32\linewidth]{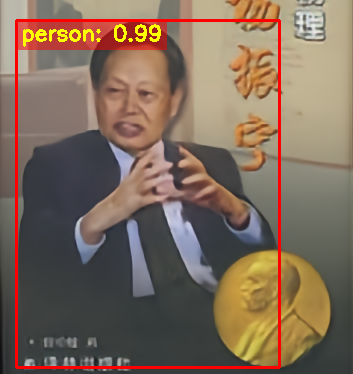}}
            \hfill
            \subfloat{\includegraphics[width=0.32\linewidth]{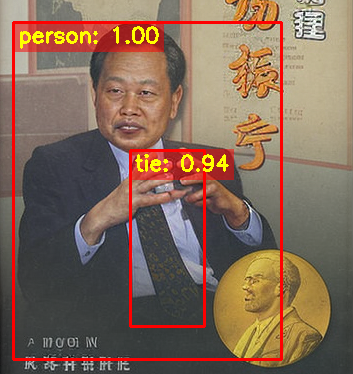}}
            \\
            \vspace{3mm}
            \addtocounter{subfigure}{-3}
            \subfloat[LQ]{\includegraphics[width=0.32\linewidth]{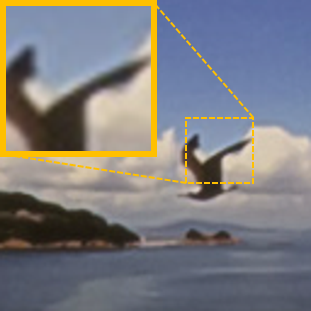}}
            \hfill
            \subfloat[SwinIR~\cite{sr_swinir}]{\includegraphics[width=0.32\linewidth]{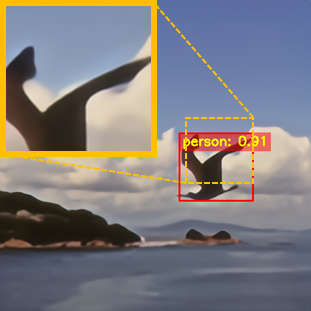}}
            \hfill
            \subfloat[\textbf{NOLA-IR}]{\includegraphics[width=0.32\linewidth]{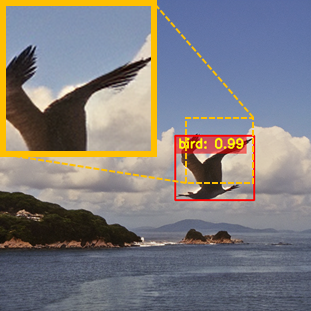}}
        \end{minipage}%
        \addtocounter{subfigure}{-3}
    }
    \caption{
    Restoration results and detection outcomes on real-world LQ inputs. Notation follows \cref{fig:vis-real-world-det} in the main paper. Rows 1--3 are from RealPhoto60~\cite{yu2024scaling}, RealLR200~\cite{wu2024seesr}, and RealSR~\cite{cai2019toward}, respectively.
    }
    \label{fig:vis-real-world-det3}
\end{figure*}

\cref{fig:vis-real-world-det1,fig:vis-real-world-det2} provide additional qualitative comparisons on real-world LQ inputs collected from (i) small patches cropped from YouTube video frames and (ii) an old-photo image crawled from the Internet.
\cref{fig:vis-real-world-det3} further provides additional results on real-world LQ datasets, including RealPhoto60~\cite{yu2024scaling}, RealLR200~\cite{wu2024seesr}, and RealSR~\cite{cai2019toward}.
All results follow the same real-world detection setting described in \cref{ssec:main-paper-real-world-det} of the main paper.

Overall, NOLA-IR restores task-relevant details more consistently and leads to more accurate detections than SwinIR, further validating strong generalization to diverse real-world degradations.

\clearpage

\clearpage

\clearpage

\section{Additional Visualization for OCR}
\label{sec:visualization_ocr}
\begin{figure*}[t!]
    \centering
    \captionsetup[subfigure]{labelfont=scriptsize, textfont=scriptsize}
    \renewcommand{\wp}{0.24}
    \renewcommand{\splitT}{0.250}
    \resizebox{1.0\linewidth}{!}{
        \begin{minipage}{\textwidth}
            \centering
            \addtocounter{subfigure}{-4}
            \subfloat{\includegraphics[width=\wp\linewidth]{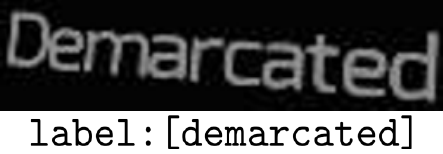}}
            \hfill
            \subfloat{\includegraphics[width=\wp\linewidth]{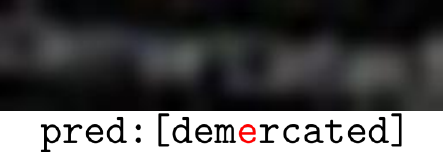}}
            \hfill
            \subfloat{\includegraphics[width=\wp\linewidth]{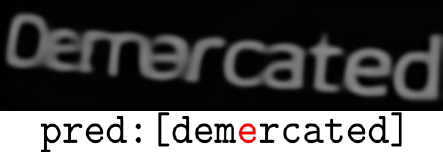}}
            \hfill
            \subfloat{\includegraphics[width=\wp\linewidth]{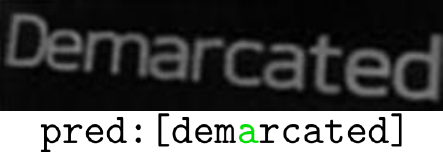}}
            \\
            \vspace{2mm}
            \addtocounter{subfigure}{-4}
            \subfloat{\includegraphics[width=\wp\linewidth]{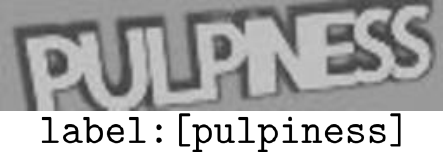}}
            \hfill
            \subfloat{\includegraphics[width=\wp\linewidth]{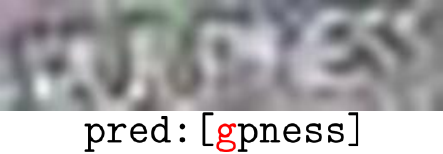}}
            \hfill
            \subfloat{\includegraphics[width=\wp\linewidth]{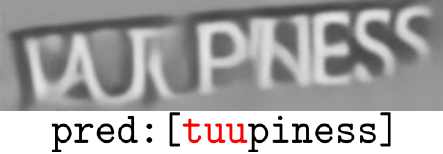}}
            \hfill
            \subfloat{\includegraphics[width=\wp\linewidth]{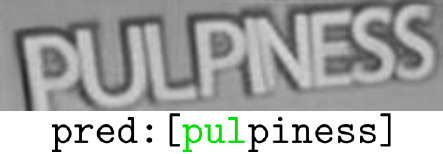}}
            \\
            \vspace{2mm}
            \addtocounter{subfigure}{-4}
            \subfloat{\includegraphics[width=\wp\linewidth]{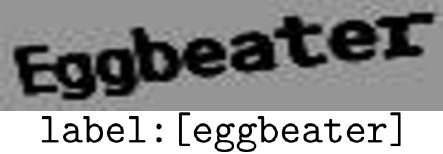}}
            \hfill
            \subfloat{\includegraphics[width=\wp\linewidth]{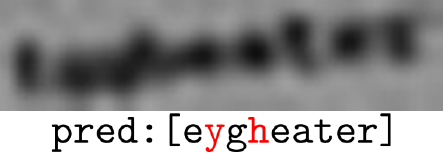}}
            \hfill
            \subfloat{\includegraphics[width=\wp\linewidth]{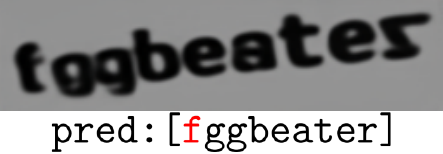}}
            \hfill
            \subfloat{\includegraphics[width=\wp\linewidth]{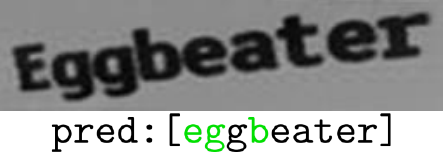}}
            \\
            \vspace{2mm}
            \addtocounter{subfigure}{-4}
            \subfloat{\includegraphics[width=\wp\linewidth]{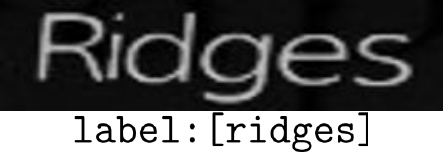}}
            \hfill
            \subfloat{\includegraphics[width=\wp\linewidth]{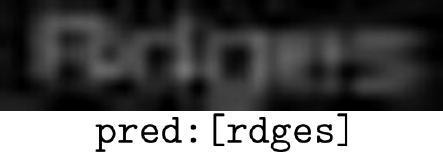}}
            \hfill
            \subfloat{\includegraphics[width=\wp\linewidth]{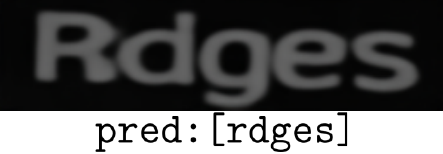}}
            \hfill
            \subfloat{\includegraphics[width=\wp\linewidth]{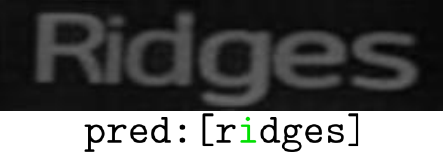}}
            \\
            \vspace{2mm}
            \addtocounter{subfigure}{-4}
            \subfloat{\includegraphics[width=\wp\linewidth]{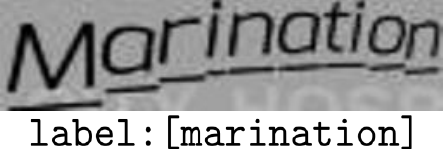}}
            \hfill
            \subfloat{\includegraphics[width=\wp\linewidth]{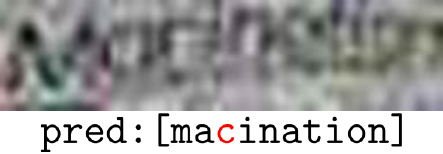}}
            \hfill
            \subfloat{\includegraphics[width=\wp\linewidth]{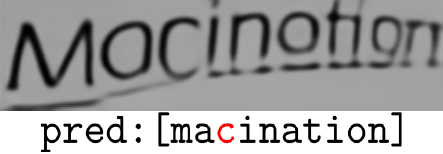}}
            \hfill
            \subfloat{\includegraphics[width=\wp\linewidth]{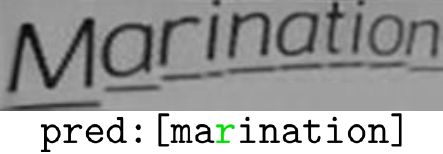}}
            \\
            \vspace{2mm}
            \addtocounter{subfigure}{-4}
            \subfloat{\includegraphics[width=\wp\linewidth]{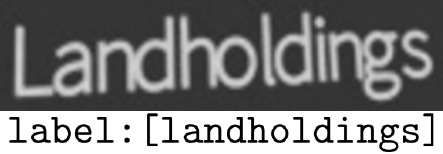}}
            \hfill
            \subfloat{\includegraphics[width=\wp\linewidth]{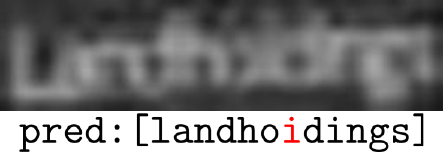}}
            \hfill
            \subfloat{\includegraphics[width=\wp\linewidth]{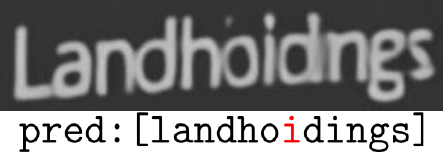}}
            \hfill
            \subfloat{\includegraphics[width=\wp\linewidth]{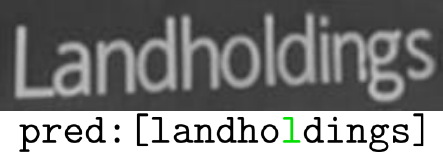}}
            \\
            \vspace{2mm}
            \addtocounter{subfigure}{-4}
            \subfloat{\includegraphics[width=\wp\linewidth]{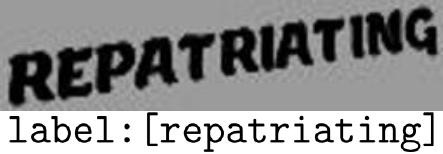}}
            \hfill
            \subfloat{\includegraphics[width=\wp\linewidth]{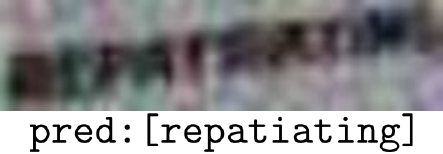}}
            \hfill
            \subfloat{\includegraphics[width=\wp\linewidth]{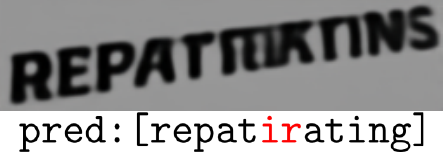}}
            \hfill
            \subfloat{\includegraphics[width=\wp\linewidth]{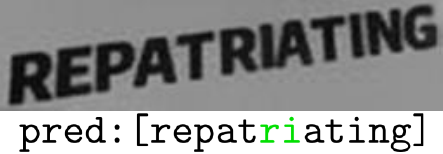}}
            \\
            \vspace{2mm}
            \addtocounter{subfigure}{-4}
            \subfloat{\includegraphics[width=\wp\linewidth]{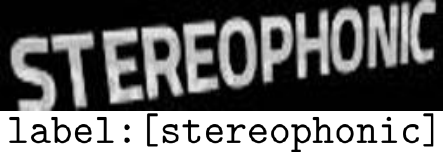}}
            \hfill
            \subfloat{\includegraphics[width=\wp\linewidth]{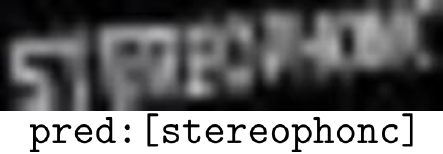}}
            \hfill
            \subfloat{\includegraphics[width=\wp\linewidth]{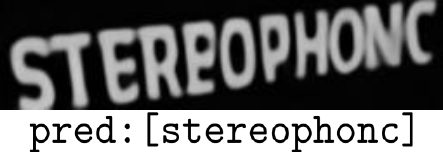}}
            \hfill
            \subfloat{\includegraphics[width=\wp\linewidth]{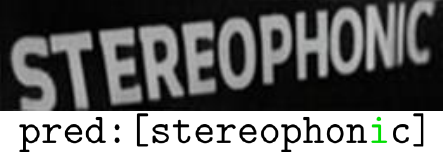}}
            \\
            \vspace{2mm}
            \addtocounter{subfigure}{-4}
            \subfloat[HQ (Ground-truth)]{\includegraphics[width=\wp\linewidth]{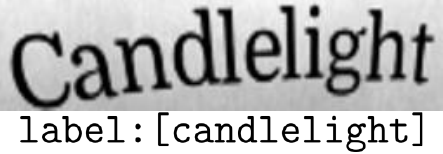}}
            \hfill
            \subfloat[LQ (Lower Bound)]{\includegraphics[width=\wp\linewidth]{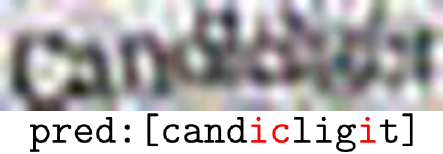}}
            \hfill
            \subfloat[SwinIR~\cite{sr_swinir}]{\includegraphics[width=\wp\linewidth]{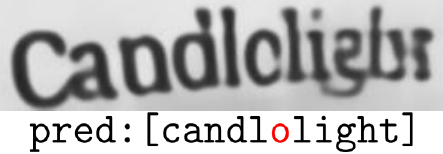}}
            \hfill
            \subfloat[\textbf{NOLA-IR}]{\includegraphics[width=\wp\linewidth]{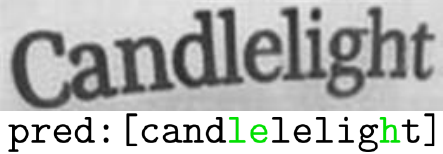}}
        \end{minipage}%
        \addtocounter{subfigure}{-8}
    }
    \caption{
        Additional qualitative OCR results on MJSynth~\cite{jaderberg2014synthetic}. Notation follows \cref{fig:vis-ocr} in the main paper.
    }
    \label{fig:additional-vis-ocr}
\end{figure*}

\cref{fig:additional-vis-ocr} provides additional qualitative OCR results on MJSynth~\cite{jaderberg2014synthetic}.
Despite the severely degraded LQ inputs, NOLA-IR effectively restores character structures and yields more accurate character recognition results than the compared methods, further supporting the generalization of our approach beyond standard high-level vision tasks.

\clearpage

\end{document}